\documentclass[10pt,twocolumn,letterpaper]{article}

\usepackage[pagenumbers]{wacv} % To force page numbers, e.g. for an arXiv version

\usepackage[toc,page]{appendix}
\usepackage{graphicx}
\usepackage{booktabs}
\usepackage{multirow}
\usepackage{subcaption}
\usepackage{amsmath}
\usepackage{amssymb}
\usepackage{datatool}
\usepackage{ dsfont }
\usepackage{placeins}
\usepackage[pagebackref,breaklinks,colorlinks]{hyperref}

\usepackage[capitalize]{cleveref}
\crefname{section}{Sec.}{Secs.}
\Crefname{section}{Section}{Sections}
\Crefname{table}{Table}{Tables}
\crefname{table}{Tab.}{Tabs.}

\begin{document}

\title{Transformer with Controlled Attention for Synchronous Motion Captioning}

\author{
    Karim Radouane$^{1,3}$ \hspace{.3cm} Sylvie Ranwez$^1$\hspace{.3cm}   Julien Lagarde$^2$\hspace{.3cm}  Andon Tchechmedjiev$^1$ \\ \vspace{1cm} 
    EuroMov DHM, IMT Mines Ales$^1$, University of Montpellier$^2$, LIPN-University Sorbonne Paris Nord$^3$
}

\maketitle

\begin{abstract}
In this paper, we address a challenging task, synchronous motion captioning, that aim to generate a language description synchronized with human motion sequences. This task pertains to numerous applications, such as aligned sign language transcription, unsupervised action segmentation and temporal grounding. Our method introduces mechanisms to control self- and cross-attention distributions of the Transformer, allowing interpretability and time-aligned text generation. We achieve this through masking strategies and structuring losses that push the model to maximize attention only on the most important frames contributing to the generation of a motion word. These constraints aim to prevent undesired mixing of information in attention maps and to provide a monotonic attention distribution across tokens. Thus, the cross attentions of tokens are used for progressive text generation in synchronization with human motion sequences. We demonstrate the superior performance of our approach through evaluation on the two available benchmark datasets, KIT-ML and HumanML3D. As visual evaluation is essential for this task, we provide a comprehensive set of animated visual illustrations in the code repository: \url{https://github.com/rd20karim/Synch-Transformer}.
\end{abstract}

\section{Introduction}
\label{sec:intro}
Motion-Language processing has garnered much interest in the computer vision community, where it has been revitalized along with increasing popularity of generative AI. In machine learning, captioning is the process of generating textual descriptions from a given input data, such as images or videos. The interest in captioning tasks stems from the need for a more efficient and effective way to understand and process visual data. Current approaches, mainly focus on often vision-based input, thus, typically relies on a combination of Convolutional Neural Networks (CNNs) and Recurrent Neural Networks (RNNs) or more recently use the Transformers \cite{Vaswani2017}. The aim is to produce detailed and human-like captions that can be used in several applications such as image and video retrieval and understanding. While captioning tasks have primarily focused on images and videos, limited research has explored motion captioning or human skeleton-based captioning \cite{Guo_2022_TM2T,Plappert2016}.

This approach generates captions for human motion based on estimated or ground-truth poses. The human skeleton offers a concise and semantically rich representation of motion, enabling better understanding and description of human activities. This task involve associating human pose sequence with close textual descriptions. The past three years have seen the emergence of larger and better quality motion-language datasets and an effervescence of ever-improving offline language to motion systems \cite{Guo_2022_CVPR}. Although such systems have been a significant focus of research \cite{Plappert2016,Petrovich2022,Guo_2022_TM2T}, there has also been an interest in motion-to-language generation \cite{Goutsu2021,Guo_2022_TM2T}, that has picked up steam with recent papers addressing synchronous motion to language generation \cite{Radouane_2023}. 

The first motion captioning architecture \cite{Radouane_2023} aiming to synchronize the generation of descriptions with human actions was based on a very simple pre-Transformer model (RNN) \cite{Cho2014} and introduced extensions to the canonical attention mechanism. Their experiments were mainly conducted on the original version of KIT-ML \cite{Plappert2016} before augmentation \cite{Guo_2022_CVPR}. While the performance exhibited was honorable, and outperformed previous offline generation systems, particularly on older and smaller datasets like KIT-ML, the emergence of larger datasets such as HumanML3D, calls for a transition to more modern architectures that have been proven to be more effective for language modelling \cite{Vaswani2017}. 

In this paper, we propose an architecture design for synchronous motion captioning based on Transformer operations. We incorporate mechanisms to control self- and cross-attention distributions, combined with structuring losses to achieve both synchronous generation along better text quality generation. We also propose masking approaches to solve mixing information problems. Subsequently, we annotate a representative subset of the test set from HumanML3D containing a more diverse range of compositional motions. This allows for an effective quantitative evaluation of the synchronization performance derived from learned attention under our proposed strategy for motion-language alignment control.

\section{Related work}
\label{sec:related_work}

In recent years, numerous motion encoders have been proposed to address the challenges of motion and text generation. Excluding studies focusing on bidirectional mapping \cite{Plappert2017,Guo_2022_TM2T}, it is evident that the field of motion generation has witnessed significant advancements, with extensive research efforts dedicated to this task \cite{Guo_2022_CVPR,Zhang2023,Ghosh2021,Petrovich2022,Chen2023}. In contrast, progress in language generation from motion has been comparatively less substantial \cite{Goutsu2021}. In this section, we will present the datasets used for both motion and language generation. Subsequently, we will discuss relevant work related to our study.

\subsection{Motion-Language Datasets}

The study of complex human movements and actions often requires the use of datasets based on motion-capture. One of the most widely used datasets is the KIT Motion Language Dataset (KIT-ML) \cite{Mandery2016}. The annotations describe the entirety of each movement, often in the form of single sentences. Recently, an updated version of the KIT-MLD dataset was introduced by augmentation \cite{Guo_2022_CVPR}, along with a much larger dataset, Human-ML3D. The Motion-Language datasets include recordings of various movement types (walking, running, waving, etc.), where the descriptions give fine-grained details specifying the body parts involved, the manner in which the motion is executed (e.g., speed).

\subsection{Motion captioning approach} 
The motion captioning task is similar to Video Captioning, where the input is a sequence of human poses instead of images. Existing motion-captioning approaches were based on recurrent neural network encoder-decoder architectures, only transitioning to using Transformer-based architectures in recent years.

\textit{RNN-based design.} A first model addressing the bidirectional generation task was proposed by \cite{Plappert2017}, using the original KIT-MLD dataset. The motion sequence is initially encoded using a stack of bidirectional RNNs to obtain a context vector $c$. This context vector is further decoded by another stack of unidirectional RNNs into a sequence of text. A similar design was used for motion generation in the reverse direction.

\textit{Transformer-based design.} More recently, both modes of generation have been addressed by \cite{Guo_2022_TM2T}. The authors proposed a transformer-based architecture to handle the generation of both text and motion. This is achieved straightforwardly by representing motion as token sequences using a codebook obtained through pretraining a VQ-VAE. Their experiments were conducted on the more recent HumanML3D dataset \cite{Guo_2022_CVPR} and augmented KIT-MLD. More recently, MotionGPT \cite{jiang2024motiongpt} involves multi-task learning: motion generation and captioning, among other tasks. The disparity in tasks prevents a fair comparison of results. However, its learning process adversely affected motion captioning, resulting in a notably low BLEU@4 score of $12.47\%$ on HumanML3D and no reported results on KIT-ML dataset for motion captioning.

\textbf{Synchronous Motion Captioning.} This task aims to provide a captioning aligned with the motion sequence represented by the human poses in time. The model learns to output a synchronized description with motion, where motion words are generated at the time of the corresponding actions. We can find some analogies with dense aligned captioning \cite{Krishna_2017_ICCV}. But the alignment is performed at the phrase level instead of the word level, and, thus, it doesn't involve progressive word generation.

\textbf{Motion primitives and description.} Synchronizing motion and language involve implicitly to localize motion primitives and their part of description in the complete sentence. This process intersect with moment retrieval that was presented as use case of \textit{text-to-motion retrieval} (\verb|TMR|) task introduced by \cite{petrovich23tmr}. \verb|TMR| model performs motion retrieval based on natural language descriptions, and shows qualitative results and initial possibilities to temporally localizing a natural language
query in a long 3D motion sequence. On the other end, synchronized captioning approaches \cite{Radouane_2023}, involve automatic unsupervised alignment, enabling a simultaneous \textit{progressive text generation} and \textit{motion segmentation}. 

\section{Methods}
\label{sec:methods}
 We aim for a motion to language system generating text synchronously while being fed a movement sequence. Like in the approach by \cite{Radouane_2023} who used a modified  NMT architecture to enable synchronicity, we propose an evolution of the Transformer architecture to achieve the same objective. In this section, we describe our contributions by going over the main components of our approach. \Cref{fig:archi_des} gives an overview on the proposed architecture design, on the left a higher level conceptual view of the interaction between the main components of the architecture, and on the right a more details schematic representation of a forward pass during inference. 

\subsection{Mixing Information in Transformer}
\label{sub_sec:mix_prob}
In the context of Neural Machine Translation (NMT), the Transformer employs a Multi-Head Attention Mechanism to learn contextualized token representations. Within each encoder layer, an input token's representation is formed as an aggregated representations of input tokens with different contributions (attention weights). This process results in context mixing \cite{quantify_mix2023}. Several studies have explored information mixing in the Transformer and its influence on predictions \cite{Schwenke2021ShowMW}, aiming to improve the use of attention for interpretability. While this mixing is effective in learning contextual representations for machine translation, it becomes misleading for interpretability analysis. The information mixing process across heads and, or even layers makes it challenging to keep track of the most relevant information used to make predictions. The increase of the number of layers makes it all the more difficult to keep track of the attention flow \cite{abnar_flow_2022} by using attention weights directly. Consequently, there are two sources of information mixing, the use of multiple Transformer Layers and the attention mechanism itself. We aim to utilize attention weights to identify the most pertinent frames that contribute to the prediction of an action word. Thus, we opt for working with a single Transformer layer. We make use of masking strategies to obtain direct information about motion time through attention, but also to construct latent \textit{compact local motion representations}. A sequence of pose frames is thus transformed into a sequence of compact motion representation which then act akin to a dictionary to retrieve the most relevant frame given a motion word query.
Additionally, introducing multiple layers in the Encoder results in an expansion of the receptive field for local motions at each layer, forming a global motion representation. Our objective is for each frame to receive information from a fixed-size window defining what is \emph{local} in the motion. This setup enables us to extract precise motion localization from the attention weights without the undesirable mixing in the information source. To prevent these behaviors, we propose masking strategies incorporated in both self and cross attention mechanisms.

Our model is fully illustrated in \Cref{fig:archi_des}. We use only one layer in the Encoder and Decoder for the reasons elicited above \ref{sub_sec:mix_prob}. Moreover, including a higher number of Transformer layers isn't documented to lead to better performance \cite{Guo_2022_TM2T} independently from interpretability.

\subsection{Masked Attention}
Let's first define the semantics of attention in the context of our task. 
The attention mechanism is based on the common concepts of Key-Query-Value, here:
\textit{Query $u_t$}: What is the most relevant local human motion information to use for the prediction of word $w_t$ ?\\
\indent \textit{Value $v_i$}: Compact local motion representation around a frame $i$.\\
\indent \textit{Key $k_i$}: Relevant key representation to learn for a value $v_i$.\\

\begin{figure*}[h]
    \begin{minipage}[b]{0.5\textwidth}
        \centering
        \begin{subfigure}[b]{\textwidth}
            \includegraphics[width=\textwidth]{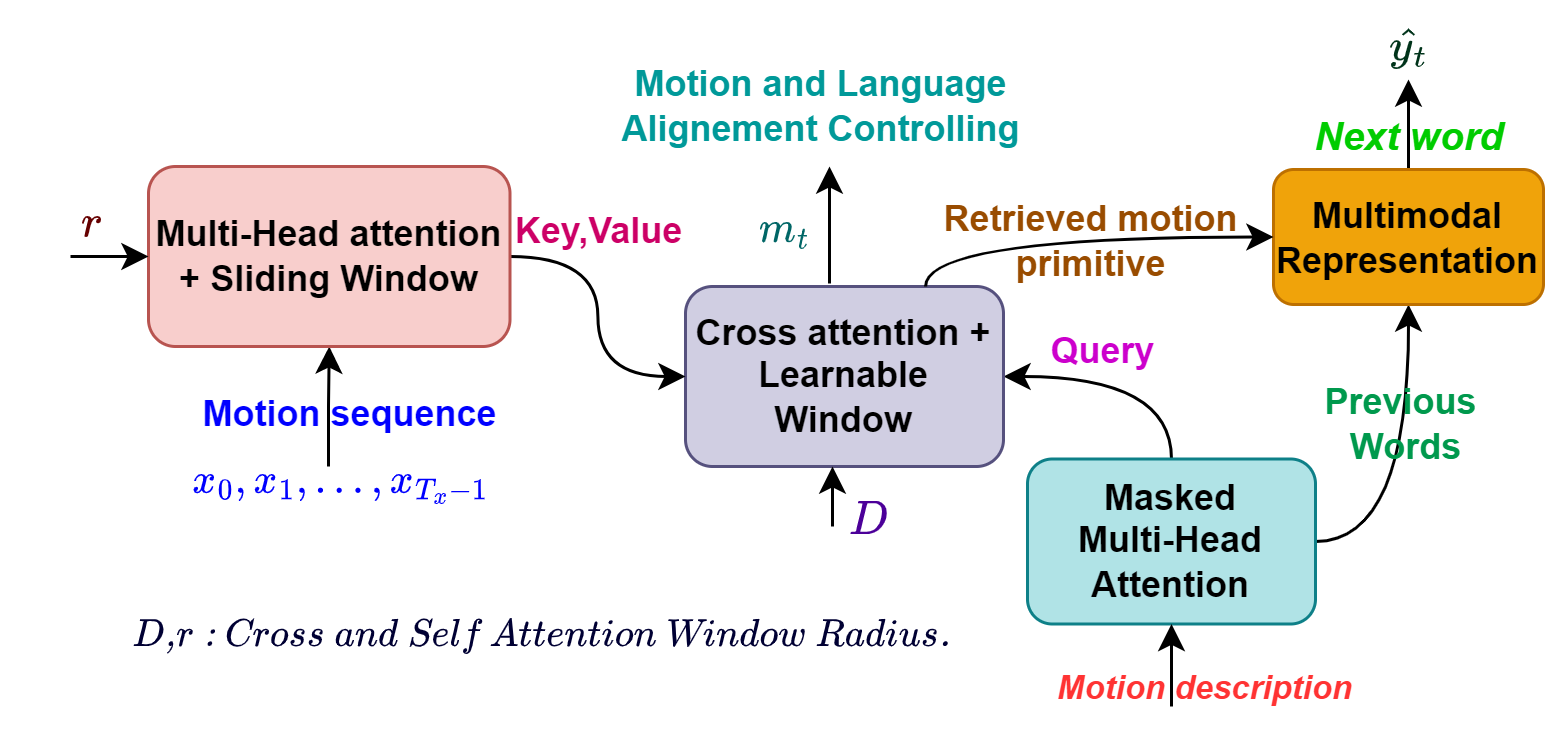}
            \caption{Information interaction in our model: A motion primitive is retrieved based on the words query through the relevant key, using weighted sum of corresponding attention scores. The alignment motion-language is controlled through  structuring losses.}
            \label{fig:concept_archi}
        \end{subfigure}
        
        \vfill 
        
        \begin{subfigure}[b]{\textwidth}
            \includegraphics[width=\textwidth]{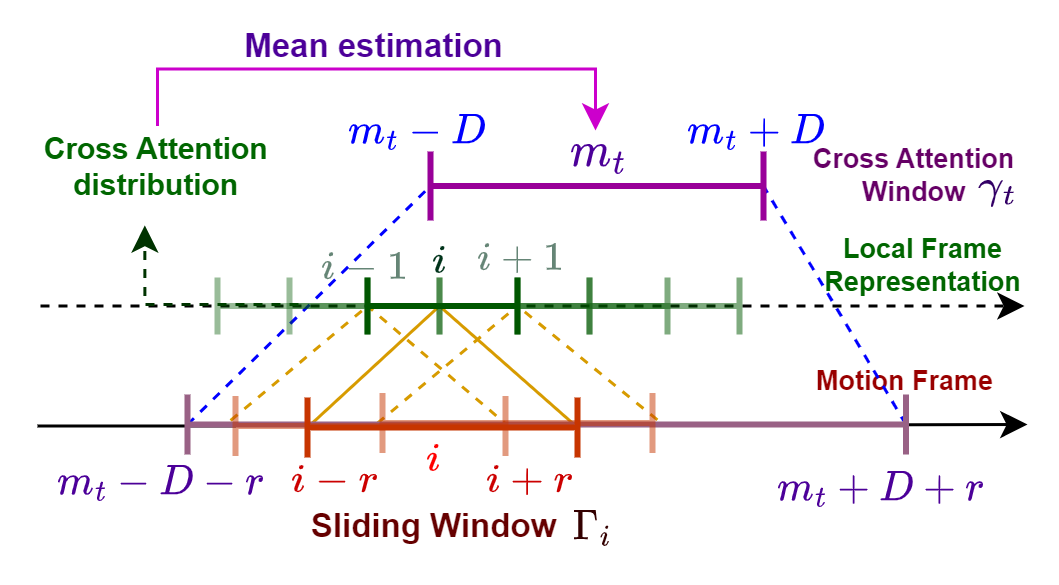}
            \caption{Receptive field of the decoder during generation spans the motion range
            $[m_t-D-r,m_t+D+r]$, where $m_t$ is the motion-word alignment position estimated from the cross attention distributions.}
            \label{fig:receptiev_field}

        \end{subfigure}
    \end{minipage}
    \hfill
    \hspace{0.001cm}
    \begin{minipage}[b]{0.48\textwidth}
        \centering
        \begin{subfigure}[b]{\textwidth}
            \includegraphics[width=\textwidth]{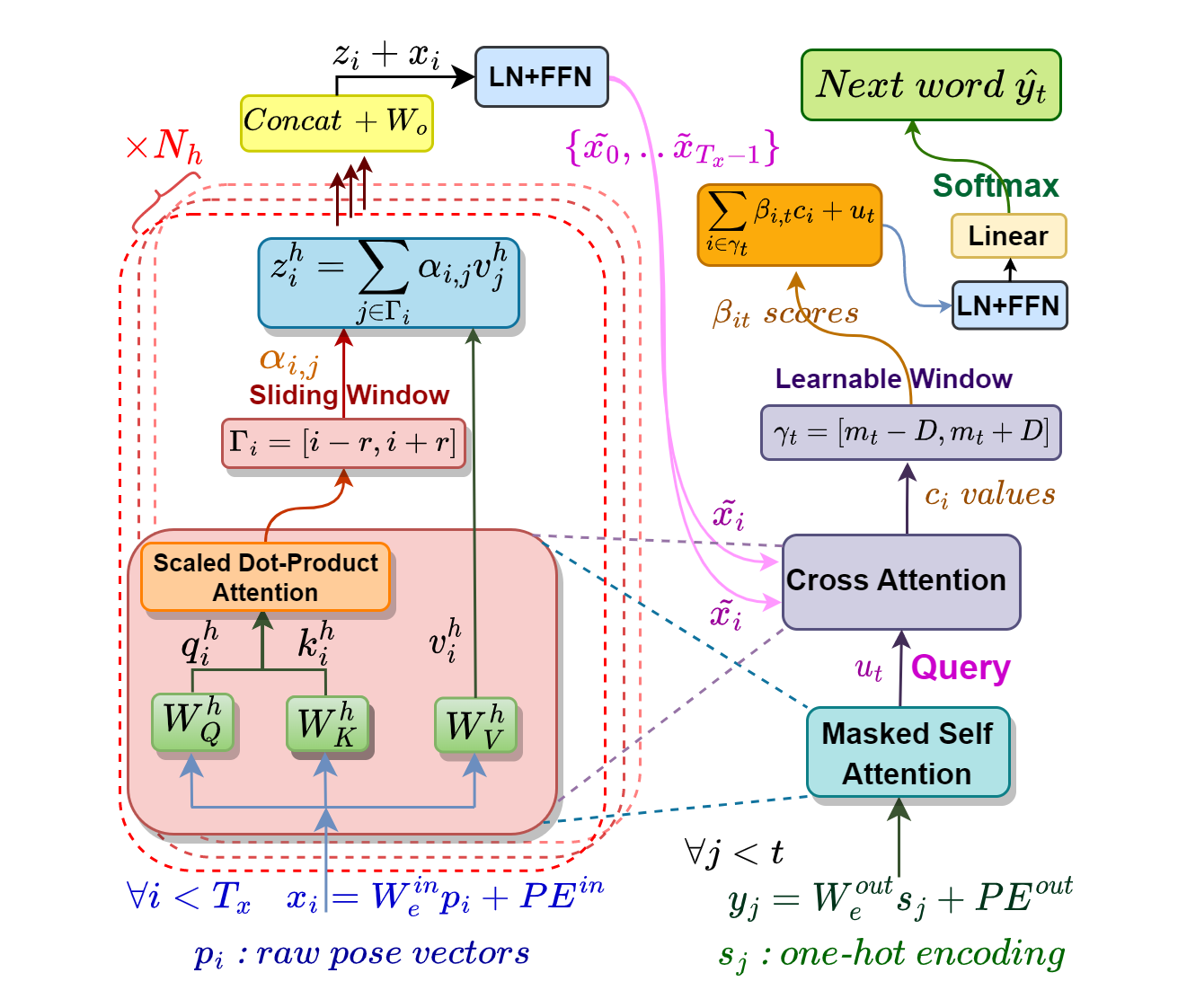}
            \caption{Our Transformer operations in inference phase: A static mask $\Gamma_i$ is incorporated in the encoder side and learnable mask $\gamma_t$ for the decoder, attention maps are controlled during training to allows the inference of synchronization time between motion and language in an unsupervised manner.}
            \label{fig:formal_archi}
        \end{subfigure}
    \end{minipage}
    \caption{Overview of general proposed framework with relevant details.}
     \label{fig:archi_des}
\end{figure*}

\textbf{Information Interaction.} As illustrated in \Cref{fig:concept_archi}, for a given query $u_t$, the goal of cross attention is to search among the provided motion keys and to retrieve the most relevant motion  values $v_j$, maximising $u_t^T\cdot k_j$ and used to predict the current word $w_t$.

\textbf{Masking Strategies.}\label{par:mask_strag} To prevent this mixing in information with long range frame communication, we propose to apply a window centered on each frame $i$ with a range of $r$ so that the new representation becomes a compact local summary of temporal information carried by frames in the range $\Gamma_i = [i-r,i+r]$. This window attention was also applied in another context of long text generation \cite{beltagy2020longformer}, referred to as \textit{sliding window}, but for different main reasons, such as computational efficiency. Masking is also incorporated in the cross-attention, as illustrated in \cref{fig:formal_archi}. In the following, we discuss the window definition for both cases in detail.

\textbf{Self Attention Window $\bf \Gamma_i$.}\label{par:slid_wind} Self attention in its original form, as proposed by \cite{Vaswani2017} lets each token attend to all other tokens. However, this results in uninformative attention weights for synchronous captioning. Here, we have a pose vector that represents the embedding of each motion frame. Using full self attention, leads to source information mixing and in turn leads to a global representation that encode information about all the actions in the sequence, while we need separate local information to localize each action involved in the human motion separately. 
Intuitively, without local masking, the representation of a frame $i$ in the next output layer may contain information about different non contiguous frames. Therefore, when the cross-attention is maximized on the final representation of frame $i$, the attention weights cannot directly be used to access the most relevant set of frames used for the current word prediction. The precise frame source of information used for the predicted motion word is lost. Moreover, including long-distance isolated frames reduces the ability of the model to learn correct local information.

\textbf{Cross Attention Window $\bf \gamma_t$.} 
\label{par:learn_wind}
We constrained attention scores to be around a learnable frame position $m_t$. This learnable value represents the center of the cross window search range $\gamma_t = [m_t-D,m_t +D]$.

\textbf{Receptive Field.} Regarding the receptive field, taking into account the two masking strategies, the query $u_t$ at step $t$ \emph{searches} in the motion frames across a window of width $L=2(D+r)$. This results in mask accumulation ranging over $[m_t -r-D,m_t+r+D]$, as illustrated in \cref{fig:receptiev_field}.

\subsection{Transformer Operations in Masking Context}

After introducing our masking strategies (Sec. \ref{par:mask_strag}), we will formulate the Transformer operations, taking into account cross- and self-attention masking.

\textbf{Multi-Head Attention.} Given a sequence of pose vectors $p_i \in \mathds{R}^c$. The pose of each frame $i$ is transformed into $x_i$ by \cref{eq:embed_pos}, where $PE$ is the common positional encoding. Then, in each head $h \in \{1,...,H\}$ in the self-attention block, ${x}_i$ is transformed into a query ${q}_i^h$, a key ${k}_i^h$, and a value ${v}_i^h$.
\begin{equation}
    \label{eq:embed_pos}
    {{x}_i = ({p}_i {W_e} + {b}_e)+PE}
\end{equation}

\begin{align}
    \label{eq:combined} 
    & &
    q_i^h &= x_i W_Q^h + b_Q^h &
    k_i^h &= x_i W_K^h + b_K^h  &
    v_i^h &= x_i W_V^h + b_V^h 
\end{align}

The context vector ${z}_i^h$ for the $i^\text{th}$ token of each attention head is then generated as a weighted sum over the transformed value vectors inside the sliding window $\Gamma_i$ (\cref{par:slid_wind}).

% \vspace{-0.3cm}
\begin{equation}
    \label{eq:context_vectors}
    {z}_i^h = \sum_{j\in \Gamma_i}\alpha_{i,j}^h {v}^h_j
\end{equation}

where $\alpha_{i,j}^h$ is the attention weight assigned to the $j^\text{th}$ frame, and computed using  \cref{eq:self_attention}. We note that scores outside the window $\Gamma_{i}$ are not considered in the soft-max operation (masked with $-\infty$).
\begin{equation}
    \label{eq:self_attention}
    \alpha^h_{i,j} = \frac{\exp{( {q^h_i}^T \cdot k^h_i  / \sqrt{d})}}{\sum_{j\in \Gamma_i} \exp{({q^h_i}^T \cdot k^h_j / \sqrt{d})}}
\end{equation}

The context vector (${z}_i \in \mathbb{R}^d$) aggregates information from each head through the ${W}_O$ projection layer \cref{eq:concat_context}. 

% \vspace{-0.3cm}
\begin{equation}
    \label{eq:concat_context}
    {z}_i = \textsc{Concat}({z}_i^1, ..., {z}_i^{N_h}){W}_O
\end{equation}

\textbf{LN + FFN.} Represent the mapping $f_{W_1^{in},W_2^{in}}:(z_i,x_i) \mapsto \tilde{x_i}$ as defined in \Cref{eq:norm_res_cont,eq:FFN,eq:normalized_FFN}.  

\begin{equation}
    \label{eq:norm_res_cont}
        {\tilde{z}}_i = {\rm LN}({z}_i + {x}_i)
\end{equation}
\vspace{-0.3cm}
\begin{equation}
    \label{eq:FFN}
    {\tilde{x}}_i = {\rm max}(0, \tilde{z}_i {W}^{in}_1 + {b}_1) {W}^{in}_2 + {b}_2 
\end{equation}
\vspace{-0.3cm}
\begin{equation}
    \label{eq:normalized_FFN}
        {\tilde{x}}_i = {\rm LN_\text{FFN}}({\tilde{x}}_i + \tilde{z}_i)
\end{equation}

Where LN is the Layer Normalization, while Feed Forward operation (FF) is given by \cref{eq:FFN}.

\textbf{Compact Local Representation.} Refers to the final motion encoding vector $\tilde{x}_i$ (\cref{eq:normalized_FFN}). Intuitively, $\tilde{x}_i$ captures local motion information centered on a frame $i$ within $\Gamma_i$.

\textbf{Cross Attention Weights.} In our cross-attention formulation we only have one attention head, and attention scores are formulated as : 

\begin{equation}
    \label{eq:cross_attention}
    \beta_{i,t} = \frac{\exp{(u_t^T \cdot k_i / \sqrt{d})}}{\sum_{j\in \gamma_t} \exp{(u_t^T \cdot k_j / \sqrt{d})}}
\end{equation}

\textbf{Retrieved Motion Primitive.} Refers to the local motion information selected as relevant for the prediction of next word $\hat{y_t}$, defined in \cref{eq:local_motion_repres}.

\begin{equation}
    \label{eq:local_motion_repres}
    r_t = \sum_{i\in \gamma_t}{\beta_{i,t}c_i}
\end{equation}

\textbf{Multimodal Representation.} Denoted as $g_t$, quantifies information about: i) previous generated words up to time $t$ given by $u_t$, and ii) local motion information $c_j$ to consider for the prediction of the next word $y_t$. Where $c_j$ is the value produced by the cross attention block  for frame $j$ (cf.\cref{fig:formal_archi}).

\begin{equation}
    \label{eq:multi_repres}
    g_t = f_{W^{out}_1,W^{out}_2}(r_t,u_t)
\end{equation}

\subsection{Transformer with Controlled Attention}

\textbf{Learnable Cross Window Center.} Given a language query $u_t$ for a motion input. Let's $A_{t}$ be the discrete random variable that associates each local motion representation around the frame $i$ to its probability $p(A_{t}=i)$ of being the most relevant information contributing to the prediction of the current word $w_t$. 
Formally, we consider the learnable center window position $m_t$ as the center of the $A_{t}$ distribution (\cref{eq:learn_mean}), where $T_x$ is the human motion length.

\begin{equation}
    \label{eq:learn_mean}
    m_t = \mathbb{E}[A_t] = \sum_{i=0}^{T_x-1} {i.p(A_{t}=i)} = \sum_{i=0}^{T_x-1}  {i.\beta_{i,t}}
\end{equation}

\textbf{Constraint on Alignment Position $\bf m_t$.}
\label{par:constraint_align}
In order to obtain synchronous generation, inspired by \cite{Radouane_2023}, we include a constraint on ${m_t}$ such that ${m_{t-1} < m_t}$ in the training loss. 
Although this constraint is language dependent and not universally true at the word level, it holds for motion words. For example, the words $\{"the", "a", "person"\}$ are not related to the monotony of frame generation, but for action words like (“walk”, “jump”), the succession  'walk'  then 'jump' happens successively in time, as results the word describing these appear successively in the human description references. The words are generated progressively with human motion evolution. Synchronous motion captioning aims to associate every set of words in the sentence describing one action to the relevant set of frames based on $m_t$ and the attention weights distribution of $A_t$.

\textbf{Initial Alignment Position.} Formally, this position is $m_0$. To  encourage the model to see the whole motion from its start, we push $m_0$ to be close to the \textit{first} motion frame and become a reference for the next learnable attention mean $m_t$, $\forall t>0$.

\textbf{Motion and Language Alignment Control.}
The model attention distributions are forced to converge toward a solution that respects the constraint ${m_{t-1} < m_t}$, $\forall t>0$ using the attention \textit{structuring losses}: 

\begin{align*}
\label{eq:align_control}
        Loss_0 &= {m_0}/{T_x} \\ 
        Loss_{m} &= \frac{1}{T_x} \sum_{t<T_x-1} \max((m_{t}+m)-m_{t+1},0)^2
\end{align*}

During training, the loss constraining monotonic positions $Loss_{m}$ will be only penalized when the constraint $m_{t}+margin \leq m_{t+1}$ is violated. We added a margin value to ensure that $m_{t+1}$ is strictly superior to $m_t$ which prevents the trivial case resulting in $m_t$ been constant for all words. This enables the \textit{attention controlling} for synchronous captioning. In all experiments, we set the margin value $m=1$.

\textbf{Training loss.} We define the global loss that can be observed as two goals of supervision mode. First, a loss term, focusing on the direct language generation. Secondly, losses focusing on attention structuring.

\begin{equation}
    Loss= Loss_{lang} + \lambda_0 Loss_0+ \lambda_m Loss_{m} 
\end{equation}

Where $(\lambda_0,\lambda_m)$ are balancing coefficients, and the language loss (\cref{eq:loss_lang}) is defined as the standard text generation objective minimizing cross entropy between the target and predicted words.

\begin{equation}
    Loss_{lang} = -\frac{1}{T_y}\sum_{j=1}^{T_y} y_{j} \log(\hat{y}_{j})
    \label{eq:loss_lang}
\end{equation}

\textbf{Attention Heads $N_h$.} While we use only one layer on both the encoder and decoder sides, multi-head attention is incorporated in both the Encoder and Decoder, except for the cross-attention which uses only one head. This choice is motivated by the necessity to capture information from different frames inside the sliding window. On the decoder side, we maintain a query that takes into consideration all previously generated words.

\begin{table*}[t]

    \centering
    \begin{tabular}{lccccccc}
    \toprule
    Dataset                  & D & r  & BLEU@1 & BLEU@4 & CIDEr & ROUGE\verb|_|L & BERTScore \\ \midrule
    \multirow{4}{*}{HML3D}    & 5 & 10 & 66.4 &     25.1 &    61.9 &      54.3 &         42.0  \\  
                             & 10 & 10  & 68.7 &     26.6 &    68.0 &      55.6 &         44.3 \\  %\cline{2-8} 
                             & 20 & 20  & \textbf{69.2} &   \bf 27.1 &    \textbf{70.3} &      \textbf{56.1} &         \textbf{45.5} \\                             
                             
                             & $\infty$ & $\infty$ & 68.9 &     26.5 &    69.0 &      56.0 &         45.0  \\

                             \midrule
    \multirow{4}{*}{KIT-ML}  & 10 & 5 & 54.3 & 21.2 & 93.7 & 54.8 & 39.0  \\ 
                            & 10 & 10  & \textbf{59.0} &  \textit{26.4} &    \textit{117.8} &      \textit{58.1} &         \textit{43.5} \\ 
                             & 20 & 20 & 57.6 & 24.4 & 116.7 & 58.1 & 44.1  \\ 
                            & $\infty$ & $\infty$ &  58.8 & \textbf{26.5} & \textbf{132.3} & \textbf{58.7} & \textbf{45.8}       \\  
                            \bottomrule

     \end{tabular}
         \caption{Controlled attention with different values for $D$ and $r$. The masking approach helped improve the NLP metrics in case of HML3D. However, these parameters have a more significant effect on our main goal of motion-language synchronization as will be demonstrated in \Cref{tab:seg_res}.}
    \label{tab:w_control}
\end{table*}

\section{Quantitative and Qualitative Results}

In our specific case, our objective extends beyond maximizing the BLEU score; we also aim to align each motion word $w_t$ with the most accurate center time of action execution. Our goal is to infer alignment information using only cross attention weights. Thus, we need to evaluate quality of both text generation and synchronicity. Given an attention distribution over frames, effective localization of an action occurs when the mean of attention weights ideally matches the center time of the action, and the start and end frames are defined by the spread of attention distribution. We will first discuss NLP metrics, qualitative analysis then evaluate synchronization.

\subsection{Ablation and Evaluation Study}
We recall that our architecture incorporates a single encoder/decoder layer Transformer. More complex designs tend to yield less interpretable attention maps and are not directly controllable. However, interpretability and attention control are crucial for inferring synchronization between motion and language in unsupervised setting. 
Consequently in our context the ablation study concern only two aspects : i) Effect of motion and language alignment controlling (structuring losses) and ii) Effect of masking approach: learnable and sliding window (more analysis in Supp.\ref{supp:ablation}).

\textbf{Hyperparameters of Attention Control.}
To enable attention control, we set $\lambda_m = 1000, \lambda_0 = 0.1$ and experiment with different values for window size, $D$ for cross attention, and $r$ for self-attention. \Cref{tab:w_control} presents quantitative results for this hyperparameter search. First, we note by $D=\infty, r=\infty$ the case where full context length is used without self- and cross-attention masking. The hidden size $d_m$ and the number of heads $N_h$ are set respectively to $128$ and $4$ for HumanML3D and to $64$ and $4$ for KIT-ML. We note that higher values of $D$ and $r$ in some cases give better results in terms of text quality (cf. \cref{tab:w_control}) but not in terms of synchronization between motion and language (cf.\cref{tab:seg_res}). Consequently, many alternative models can yield good or equivalent solutions in terms of text quality generation, but not all lead to good synchronization.

\textbf{Comparison with SOTAs.}
Although our primary objective goes beyond merely enhancing the quality of the generated text, for comparison, we present the standard text generation metrics in \Cref{table:comp_sota} based solely on text quality generation. On KIT-ML, our model significantly outperforms the \verb|TM2T| model which is also Transformer-based model but with $3$ layers in the Encoder and Decoder. In contrast, our model employs only one layer with fewer parameters and does not utilize beam searching, while achieving synchronous captioning. In comparison to the model \verb|MLP+GRU|, we achieve significantly better results than SOTA both on KIT-ML (SOTA + 1.1\% BLEU4, -0.1\% ROUGE, +6.6 CIDEr, +3.7\% BERTScore), and on HumanML3D (SOTA + 3.7\% BLEU4, +2.3\% ROUGE, -2.2 CIDEr, +8.3\% BERTScore).

\begin{table*}[t] 

\centering
{
    \begin{tabular}{cccccccc}
    \toprule
     Dataset  &  Model  & BLEU@1      &  BLEU@4 & ROUGE-L & CIDEr & Bertscore\\ \midrule
    \multirow{9}{*}{\bf KIT-ML} & RAEs \cite{yamada2018paired}     &  30.6       &  0.10 & 25.7 & 8.00  & 0.40 \\ 
                            ~       & Seq2Seq(Att)                  &  34.3      &  9.30 & 36.3 & 37.3   & 5.30 \\ 
                        ~       & SeqGAN \cite{Goutsu2021}         &  3.12      & 5.20 & 32.4 & 29.5    & 2.20 \\ 
                            ~       & TM2T w/o MT \cite{Guo_2022_TM2T}                  &  42.8       &  14.7 & 39.9 & 60.1  & 18.9 \\ 
                        ~       & TM2T \cite{Guo_2022_TM2T}    &  46.7      & 18.4 & 44.2 & 79.5     & 23.0 \\
                         ~       & MLP+GRU \cite{Radouane_2023}      & 56.8 & 25.4 & \textbf{58.8}  & 125.7 & 42.1 \\
                         ~  & [Spat+Adapt](2,3) \cite{radouane2024guided}    & 58.4 & 24.7  & 57.8 & 106.2 & 41.3 \\ \cline{2-7}
                             ~ & \textbf{Ours}               & \bf 58.8&  \bf 26.5  & 58.7 & \textbf{132.3} & \bf  45.8 \\  

    \midrule
    \multirow{9}{*}{\bf HML3D} & RAEs \cite{yamada2018paired}       &  33.3       &  10.2 & 37.5 & 22.1 & 10.7 \\ 
                            ~   & Seq2Seq(Att)                      &  51.8      &  17.9 & 46.4 & 58.4  & 29.1 \\ 
                        ~       & SeqGAN \cite{Goutsu2021}          &  47.8      & 13.5 & 39.2 & 50.2   & 23.4 \\ 
                            ~   & TM2T w/o MT \cite{Goutsu2021}                     &  59.5       &  21.2 & 47.8 & 68.3 & 34.9 \\ 
                        ~       & TM2T \cite{Guo_2022_TM2T}   &  61.7      & 22.3 & 49.2 & \textbf{72.5}   & 37.8 \\  
                       ~        & MLP+GRU \cite{Radouane_2023}     &  67.0 & 23.4 & 53.8  &  53.7 & 37.2 \\ 
                      ~         & [Adapt](0,3) \cite{radouane2024guided}  & 67.9 & \textbf{25.5}  & 54.7 & 64.6 & 43.2  \\ \cline{2-7}
                    ~ & \textbf{Ours}         &  \bf 69.2 &   \bf 27.1  &      \bf 56.1 &     70.3  &  \bf   45.5 \\ 
          \bottomrule
    
    \end{tabular}
}
\caption{Text generation performance conditioned on human pose motion sequence. Beyond our motion-language synchronization goal, our approach performs significantly better across different NLP metrics.}
\label{table:comp_sota}
\end{table*}

\subsection{Qualitative analysis}
In this part we discuss qualitative results at the level of attention maps and human motion sequences frozen in time.

\begin{figure}[h]
    \centering
    \begin{subfigure}{\columnwidth}
        \centering
        \includegraphics[width=\textwidth]{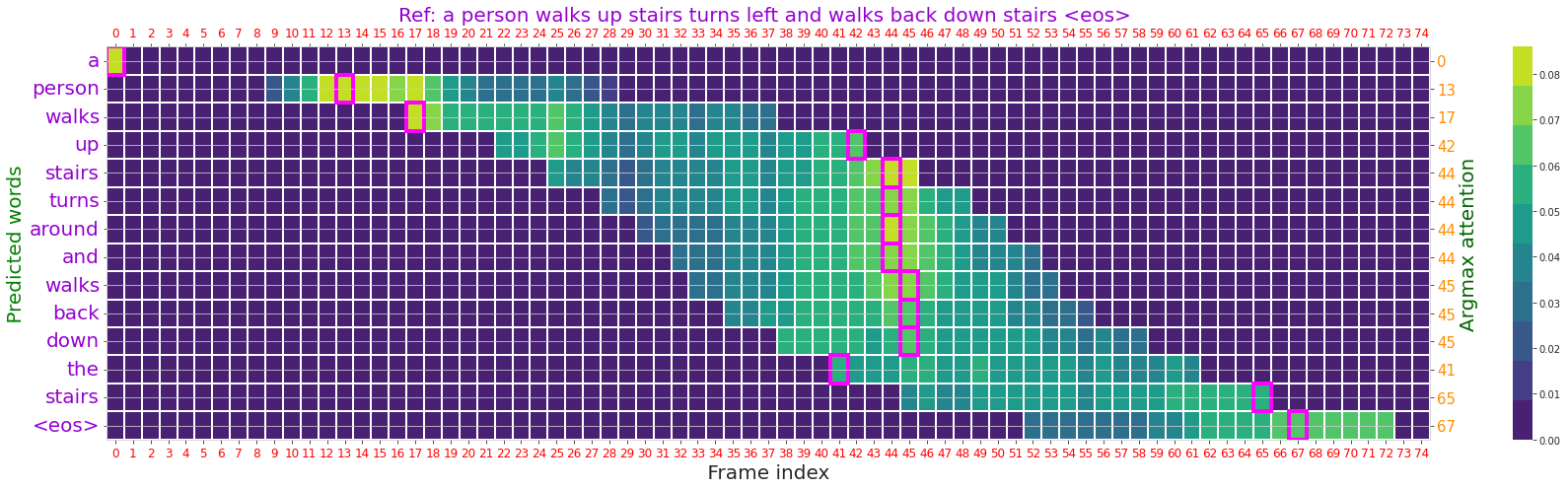}
        \caption{walk stairs [10,40], turn [41,59], walk down stairs [60,74]}
        \label{fig:walk_turn_walk}
    \end{subfigure} \vfill
    \begin{subfigure}{\columnwidth}
        \centering
        \includegraphics[width=\textwidth]{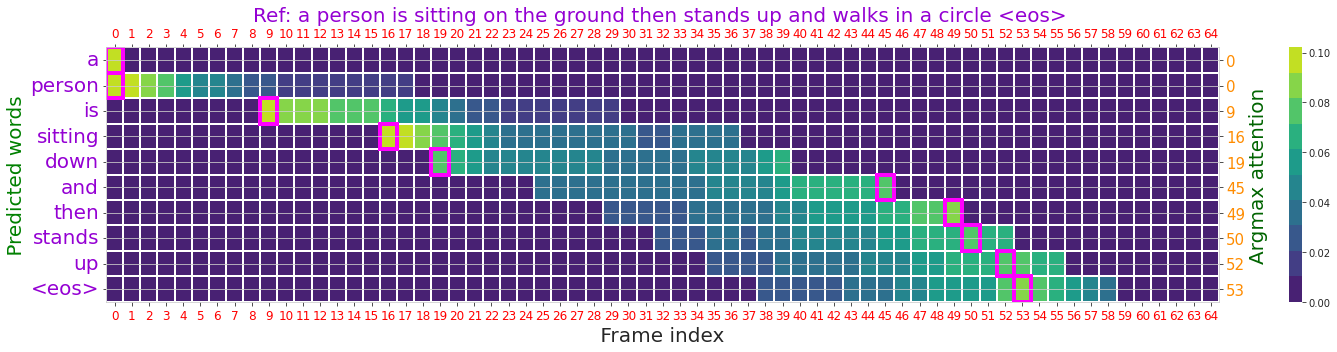}
        \caption{sitting [0,40], stand-up [41,60].}
        \label{fig:sit_stand}
    \end{subfigure}  \vfill
    \begin{subfigure}{\columnwidth}
        \centering
        \includegraphics[width=\textwidth]{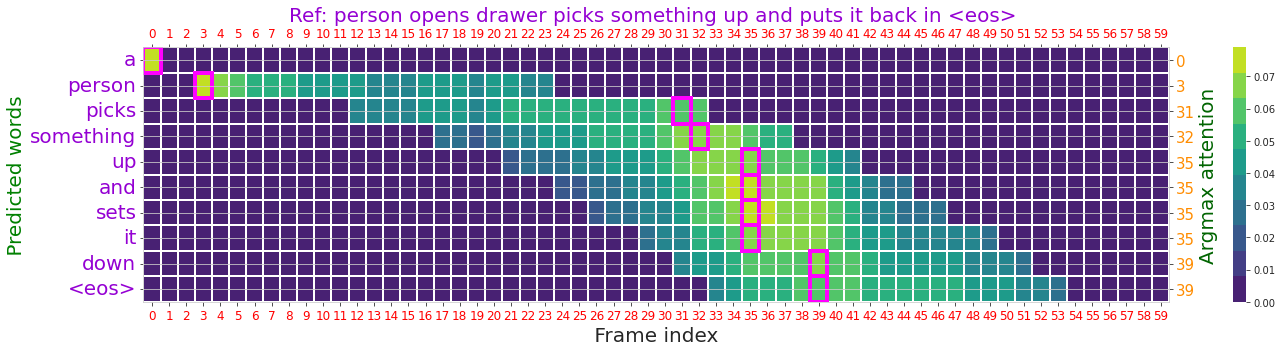}
        \caption{pick something [10,20], put it down [28,50]}
        \label{fig:pick_put}
    \end{subfigure}
    \caption{Cross attention map of compositional motions with corresponding frame range of each action (D=r=10). Across multiple examples, we observe that the attention distribution of motion words consistently falls within the indicated motion range for each specific action.}
    \label{fig:cross_att_comps}
\end{figure}

\textbf{Cross Attention Maps.} Examples of compositional motions are shown in \Cref{fig:cross_att_comps} with corresponding motion ranges. The violet rectangles represents the position of maximum attention. Each word is generated at it's corresponding position. In \cref{fig:walk_turn_walk}, considering the motion words, the spread of attention for the phrase \textit{walks up stairs} is in the range [17, 45], as compared to the manuel observation [10, 40]. The subject turns at Frame $45$, where the predicted attention for the word \textit{turn} is maximal at the frame $44$. Similar analyses could be conducted on other samples (\cref{fig:sit_stand,fig:pick_put}). However, the evaluation remains subjective, specifically in terms of defining the start/end of each action. To address this limitation, the \textit{Intersection over Prediction} (IoP) and \textit{Element of} metrics were proposed by \cite{Radouane_2023}.

\begin{figure}[t]
    \centering
    \begin{subfigure}{.95\columnwidth}
        \centering
        \includegraphics[width=\textwidth]{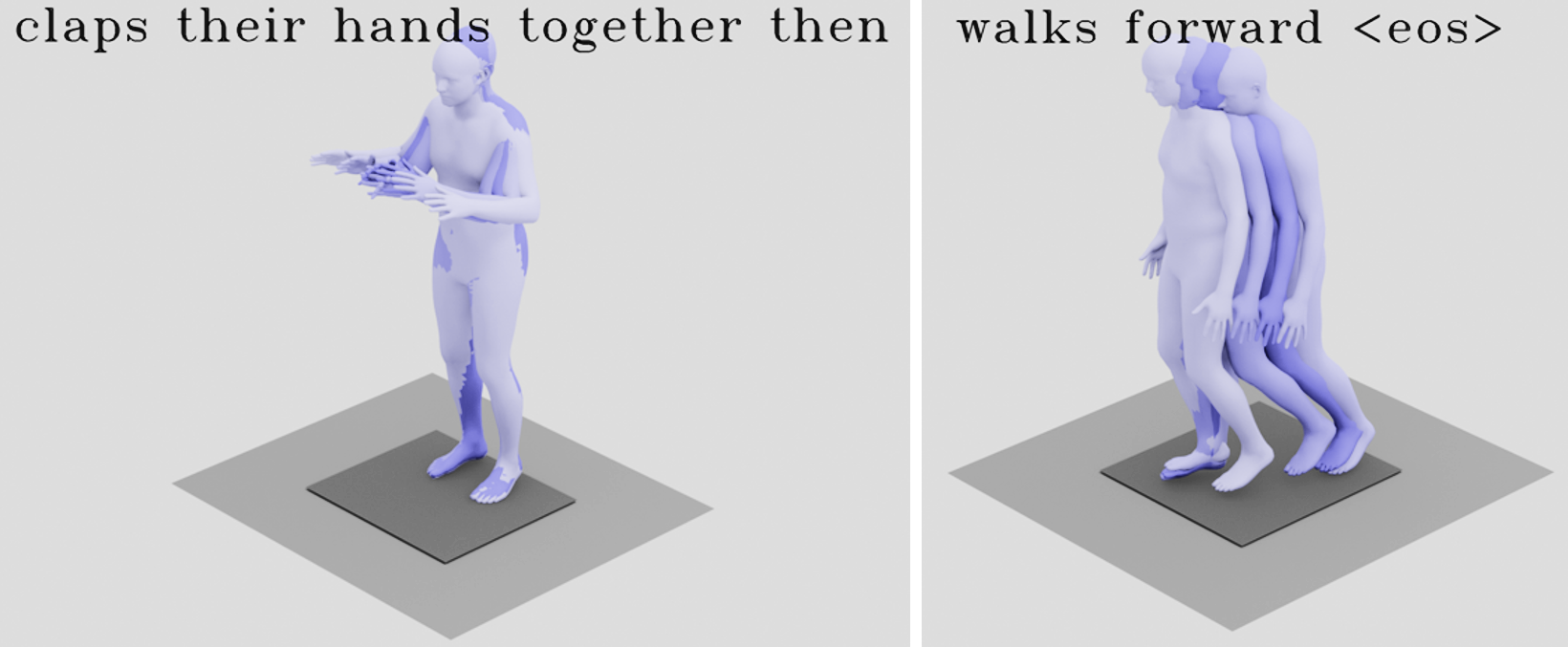}
        \caption{claps then walks.}
        \label{fig:clap_walk}
    \end{subfigure} \vfill
    \begin{subfigure}{.95\columnwidth}
        \centering
         \includegraphics[width=\textwidth]{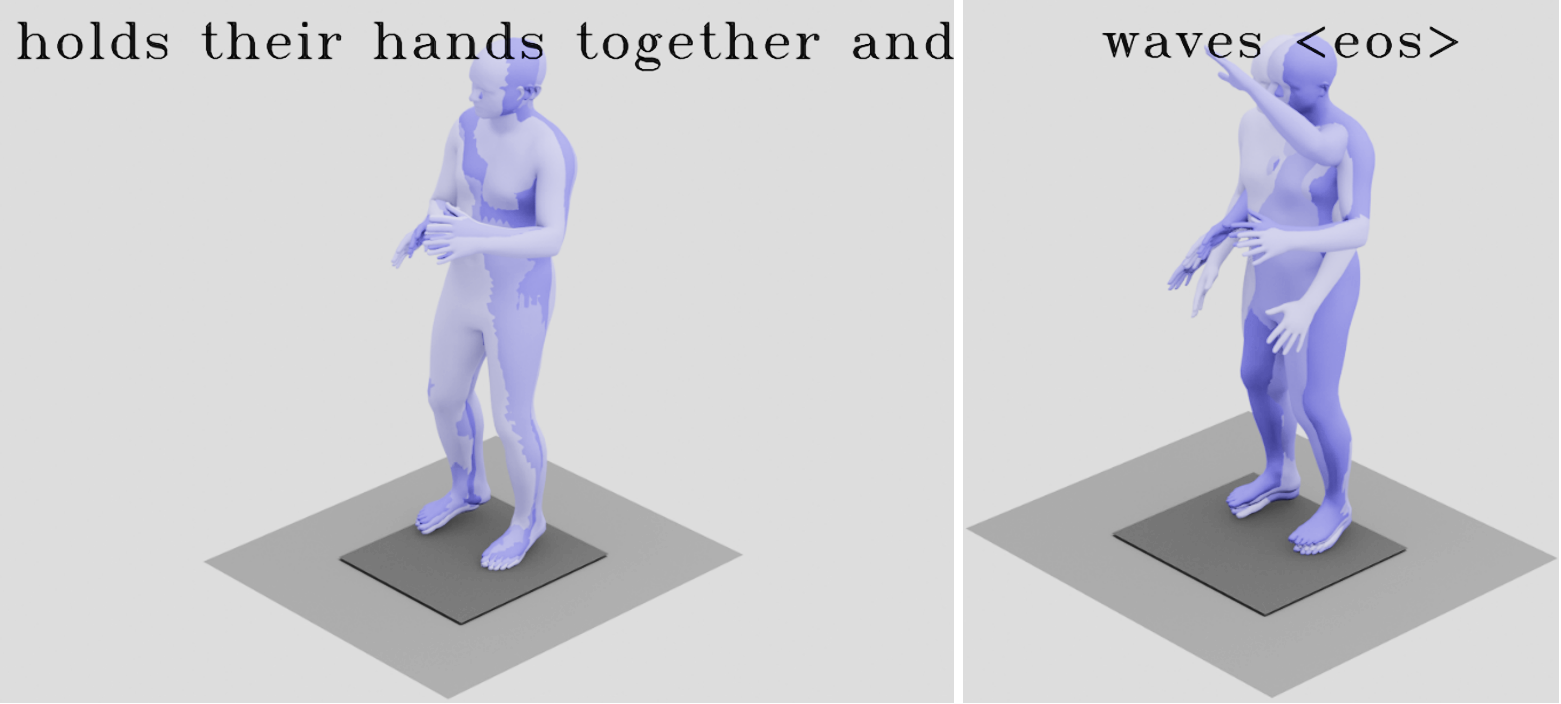}
        \caption{holds hands then waves.}
        \label{fig:left_then_right}
        \includegraphics[width=\textwidth]{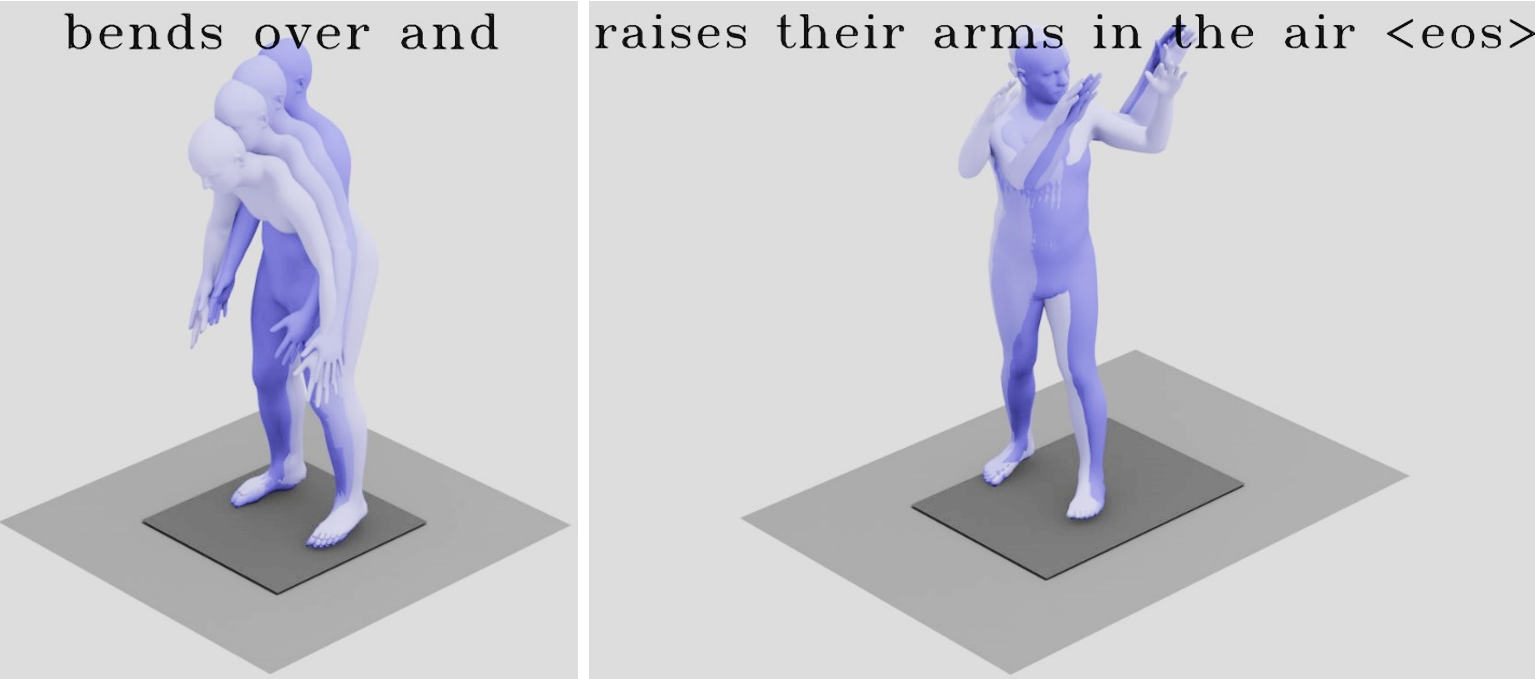}
        \caption{bends then raises arms.}
        \label{fig:bends_raises}
    \end{subfigure}
    \caption{Frozen motion with 4 keyframes of higher attention corresponding to the language segment.}
    \label{fig:frozen_samples_}
\end{figure}

\begin{figure}[t]
    \centering
    \begin{subfigure}{\columnwidth}
        \centering
        \includegraphics[width=\textwidth]{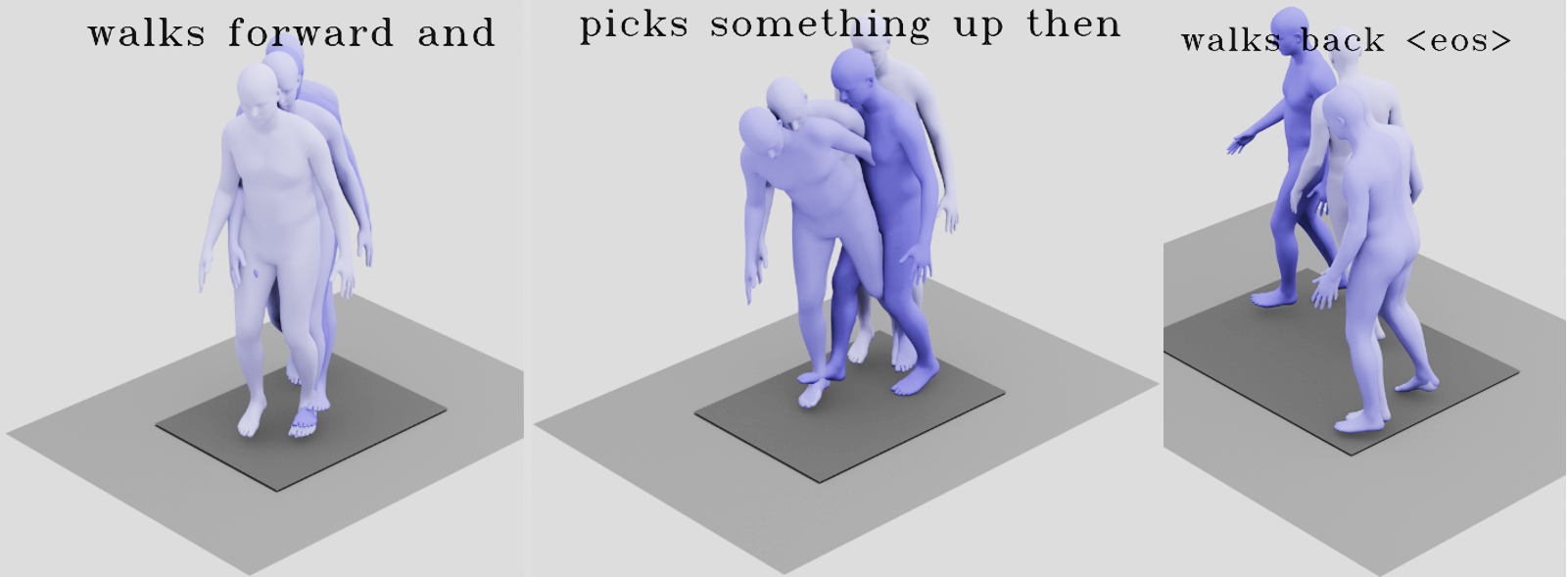}        
        \label{fig:sub_fig_label}
        \caption{a person walks forward picks something up with their right hand and walks back.}
    \end{subfigure}\vfill
        \begin{subfigure}{\columnwidth}
        \centering
        \includegraphics[width=\textwidth]{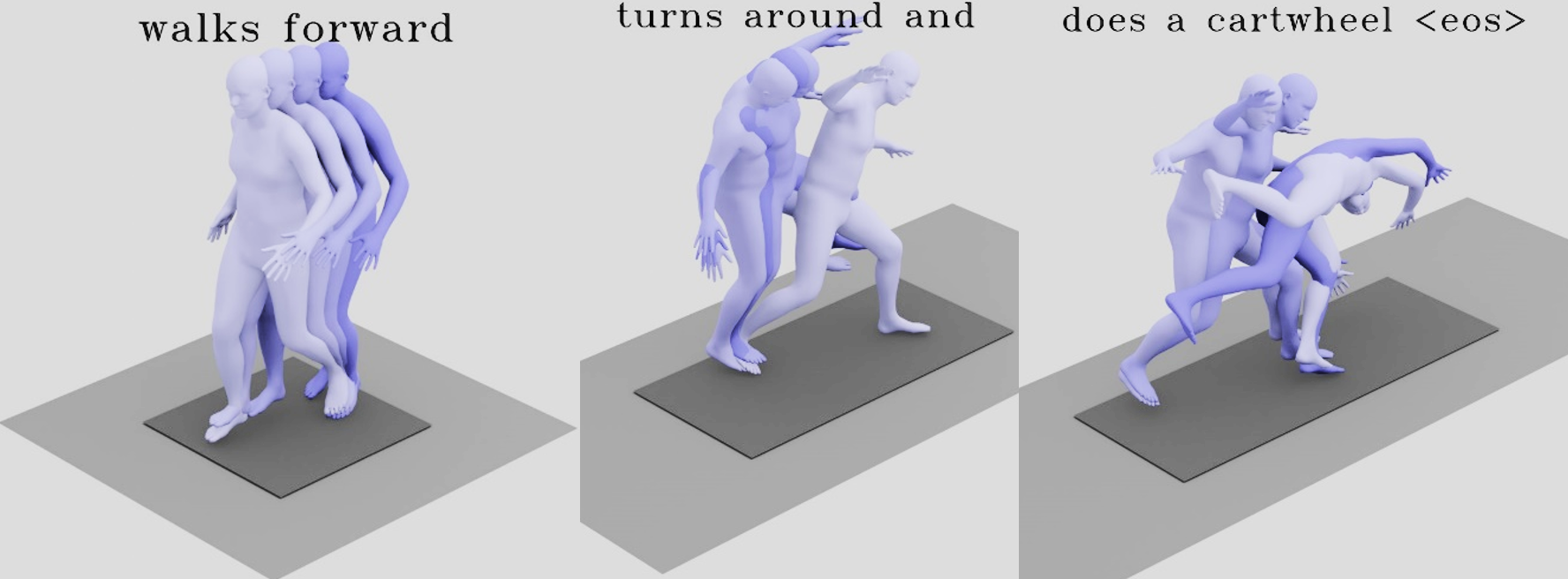}        
        \label{fig:sub_fig_label}
        \caption{a person walks forward turns around and does a cartwheel.}
    \end{subfigure}
    
    \caption{Decomposition of motions and associated descriptions (Animations in the code repository, other visualizations in Supp.\ref{supp:more_vizu}).}
    \label{fig:frozen_2}
\end{figure}

\textbf{Motion Frozen in Time.} We use static visualizations to illustrate, at a single point in time, the association of motion words with the frames receiving maximum attention. \Cref{fig:frozen_samples_} illustrates motion phrases and their sequence of frames at maximum attention. More illustrations are given in \Cref{fig:frozen_2}. However, this static visualization still have their inherent limitation, so we include animations in the code page\footnote{\url{https://github.com/rd20karim/Synch-Transformer}}.

\subsection{Evaluating word-motion synchronicity}

In this section, to quantify the synchronization between a human pose sequence and the corresponding motion-description words we use the metrics \textit{IoP}, \textit{IoU} and \textit{Element of} proposed in \cite{Radouane_2023}. However, the subjective nature of captioning process and time labeling make it difficult to consider exclusively metric values, as results, it remains very challenging and serves as quantitative complementary measure to visual animations. \textit{These animations of synchronous text generation can be found in our code repository.}

\textbf{Annotation.} First, we annotate a representative subset of the test set from Human-ML3D, which is richer in diverse compositional motions. We select samples from different actions featuring compositional motions, each containing at least two actions, to ensure an effective evaluation of synchronicity.

\textbf{Metrics.} We assess the alignment between a primitive human motion and its description based on motion words. We identify the frame time with maximum attention given to a motion and word, and then test whether the frame time falls within the motion action range (utilizing the \textit{Element of} method). Effective synchronization involves outputting each motion word during its corresponding motion execution. In contrast, IoP and IoU metrics primarily gauge the accuracy of localizing the start/end of each action. These metrics were introduced in detail by \cite{Radouane_2023}. Observing the results in \Cref{tab:seg_res}, we can conclude that $D=r=10$ provides the best tradeoff between the quality of text generation and synchronization.

\begin{table}[ht]

    \centering
   \begin{tabular}{cccccc}
        \toprule
        \textbf{D} & \textbf{r} & \textbf{IoU}   &  \textbf{IoP}    & \textbf{Element of } & \textbf{BLEU@4}  \\ \midrule
        20 & 20 & \textbf{51.35}&  60.55& 71.55 &  \textbf{27.1}\\ 
        \textbf{10} & \textbf{10} & \textit{46.40}&  \textbf{67.96}& \textbf{78.48} & \textit{26.6}\\ 
        5 & 10 & 45.23&  \textit{62.40} & \textit{75.62} & 25.1 \\ 
        $\infty$ & $\infty$ & 39.93 & 39.96 & 46.98 & 26.5 \\  \bottomrule
    \end{tabular}
    \caption{Synchronization scores for different $D$ and $r$ values show that these parameters have a more significant effect on action localization (IoP/IoU) and synchronicity (Element of). Our masking approach with $(D=r=10)$ prevents the mixing of information from different actions, enabling a better attention-based localization of action time compared to $(D=r=\infty)$, despite having slightly the same BLEU score.} 
    \label{tab:seg_res}
\end{table}

\section{Applications}

In recent times, substantial advancements have been achieved in the domain of sign language research, focusing on various specific objectives, including \textit{alignment} \cite{bull21bslalign}, \textit{temporal localization} \cite{varol21_bslattend}, and \textit{sign spotting} \cite{momeni20_spotting}. In line with these efforts, a related approach in this field, proposed by \cite{varol21_bslattend}, also uses attention scores to identify and segment signs in continuous video. 

\textbf{Aligned sign language translation.} Building an automated sign language translator with alignment information involves associating sign segments with their corresponding language segments. This task implicitly aims to link a sequence of upper-body pose movements to words, and the proposed approach can be employed to create techniques for achieving this alignment in an unsupervised manner.

\textbf{Temporal action localization.} For skeleton based action localization within a continuous stream, this task could be formulated as mapping a sequence of poses to a sequence of actions. Utilizing cross attention weights in this scenario enables the unsupervised inference of action start/end times, eliminating the necessity for labeled action time data. When time annotations are accessible, they can guide the supervision of temporal weight distribution, thereby improving the accuracy of action localization and providing more interpretable attention maps.

\section{Conclusion}
In the future, we may explore more advanced methods for local motion representation, including the incorporation of multiple heads in cross-attention. However, improving synchronous captioning remains challenging, as it requires tracking the interaction between different attention weights sources. We plan to leverage existing attention aggregation methods. Furthermore, it's worth noting that the presented methodologies hold promise for application in various scenarios beyond our current task, such as alignment for sign language translation and unsupervised action segmentation. We believe that taking steps towards controlling attention weights can lead to more explainable solutions, especially in resolving multiple tasks in unsupervised settings.

\onecolumn

%%%%%%%%% REFERENCES
{\small
\bibliographystyle{ieee_fullname}
\bibliography{main}

\begin{thebibliography}{10}\itemsep=-1pt

\bibitem{abnar_flow_2022}
Samira Abnar and Willem Zuidema.
\newblock Quantifying attention flow in transformers.
\newblock In {\em Proceedings of the 58th Annual Meeting of the Association for Computational Linguistics}, pages 4190--4197, Online, July 2020. Association for Computational Linguistics.

\bibitem{beltagy2020longformer}
Iz Beltagy, Matthew~E. Peters, and Arman Cohan.
\newblock Longformer: The long-document transformer.
\newblock 2020.

\bibitem{bull21bslalign}
Hannah Bull, Triantafyllos Afouras, G{\"u}l Varol, Samuel Albanie, Liliane Momeni, and Andrew Zisserman.
\newblock Aligning subtitles in sign language videos.
\newblock In {\em ICCV}, 2021.

\bibitem{Chen2023}
Xin Chen, Biao Jiang, Wen Liu, Zilong Huang, Bin Fu, Tao Chen, and Gang Yu.
\newblock Executing your commands via motion diffusion in latent space.
\newblock In {\em Proceedings of the IEEE/CVF Conference on Computer Vision and Pattern Recognition}, pages 18000--18010, 2023.

\bibitem{Cho2014}
Kyunghyun Cho, Bart van Merri{\"e}nboer, Caglar Gulcehre, Dzmitry Bahdanau, Fethi Bougares, Holger Schwenk, and Yoshua Bengio.
\newblock Learning phrase representations using {RNN} encoder{--}decoder for statistical machine translation.
\newblock In {\em Proceedings of the 2014 Conference on Empirical Methods in Natural Language Processing ({EMNLP})}, pages 1724--1734, Doha, Qatar, Oct. 2014. Association for Computational Linguistics.

\bibitem{Ghosh2021}
Anindita Ghosh, Noshaba Cheema, Cennet Oguz, Christian Theobalt, and Philipp Slusallek.
\newblock Synthesis of compositional animations from textual descriptions.
\newblock In {\em 2021 IEEE/CVF International Conference on Computer Vision (ICCV)}, pages 1376--1386, 2021.

\bibitem{Goutsu2021}
Yusuke Goutsu and Tetsunari Inamura.
\newblock Linguistic descriptions of human motion with generative adversarial seq2seq learning.
\newblock In {\em Proceedings of the 2021 IEEE International Conference on Robotics and Automation (ICRA)}, pages 4281--4287. IEEE, 2021.

\bibitem{Guo_2022_CVPR}
Chuan Guo, Shihao Zou, Xinxin Zuo, Sen Wang, Wei Ji, Xingyu Li, and Li Cheng.
\newblock Generating diverse and natural 3d human motions from text.
\newblock In {\em Proceedings of the IEEE/CVF Conference on Computer Vision and Pattern Recognition (CVPR)}, pages 5152--5161, June 2022.

\bibitem{Guo_2022_TM2T}
Chuan Guo, Xinxin Zuo, Sen Wang, and Li Cheng.
\newblock Tm2t: Stochastic and tokenized modeling for the reciprocal generation of 3d human motions and texts.
\newblock In Shai Avidan, Gabriel Brostow, Moustapha Ciss{\'e}, Giovanni~Maria Farinella, and Tal Hassner, editors, {\em Computer Vision -- ECCV 2022}, pages 580--597, Cham, 2022. Springer Nature Switzerland.

\bibitem{jiang2024motiongpt}
Biao Jiang, Xin Chen, Wen Liu, Jingyi Yu, Gang Yu, and Tao Chen.
\newblock Motiongpt: Human motion as a foreign language.
\newblock {\em Advances in Neural Information Processing Systems}, 36, 2024.

\bibitem{Krishna_2017_ICCV}
Ranjay Krishna, Kenji Hata, Frederic Ren, Li Fei-Fei, and Juan Carlos~Niebles.
\newblock Dense-captioning events in videos.
\newblock In {\em Proceedings of the IEEE International Conference on Computer Vision (ICCV)}, Oct 2017.

\bibitem{Mandery2016}
Christian Mandery, Ömer Terlemez, Martin Do, Nikolaus Vahrenkamp, and Tamim Asfour.
\newblock Unifying representations and large-scale whole-body motion databases for studying human motion.
\newblock {\em IEEE Transactions on Robotics}, 32:796--809, 8 2016.

\bibitem{quantify_mix2023}
Hosein Mohebbi, Willem Zuidema, Grzegorz Chrupa{\l}a, and Afra Alishahi.
\newblock Quantifying context mixing in transformers.
\newblock In {\em Proceedings of the 17th Conference of the European Chapter of the Association for Computational Linguistics}, pages 3378--3400, Dubrovnik, Croatia, May 2023. Association for Computational Linguistics.

\bibitem{momeni20_spotting}
Liliane Momeni, G{\"u}l Varol, Samuel Albanie, Triantafyllos Afouras, and Andrew Zisserman.
\newblock Watch, read and lookup: learning to spot signs from multiple supervisors.
\newblock In {\em ACCV}, 2020.

\bibitem{Petrovich2022}
Mathis Petrovich, Michael~J. Black, and G{\"u}l Varol.
\newblock {TEMOS}: Generating diverse human motions from textual descriptions.
\newblock In {\em European Conference on Computer Vision ({ECCV})}, 2022.

\bibitem{petrovich23tmr}
Mathis Petrovich, Michael~J. Black, and G{\"u}l Varol.
\newblock {TMR}: Text-to-motion retrieval using contrastive {3D} human motion synthesis.
\newblock In {\em ICCV}, 2023.

\bibitem{Plappert2016}
Matthias Plappert, Christian Mandery, and Tamim Asfour.
\newblock The {KIT} motion-language dataset.
\newblock {\em Big Data}, 4(4):236--252, dec 2016.

\bibitem{Plappert2017}
Matthias Plappert, Christian Mandery, and Tamim Asfour.
\newblock Learning a bidirectional mapping between human whole-body motion and natural language using deep recurrent neural networks.
\newblock {\em Robotics and Autonomous Systems}, 109:13--26, 5 2017.

\bibitem{radouane2024guided}
Karim Radouane, Julien Lagarde, Sylvie Ranwez, and Andon Tchechmedjiev.
\newblock Guided attention for interpretable motion captioning.
\newblock In {\em Proceedings of the 35th British Machine Vision Conference}, 2024.

\bibitem{Radouane_2023}
Karim Radouane, Andon Tchechmedjiev, Julien Lagarde, and Sylvie Ranwez.
\newblock Motion2language, unsupervised learning of synchronized semantic motion segmentation.
\newblock {\em Neural Computing and Applications}, 36(8):4401–4420, Dec. 2023.

\bibitem{Schwenke2021ShowMW}
Leonid Schwenke and Martin Atzmueller.
\newblock Show me what you're looking for visualizing abstracted transformer attention for enhancing their local interpretability on time series data.
\newblock {\em The International FLAIRS Conference Proceedings}, 34, 2021.

\bibitem{varol21_bslattend}
G{\"u}l Varol, Liliane Momeni, Samuel Albanie, Triantafyllos Afouras, and Andrew Zisserman.
\newblock Read and attend: Temporal localisation in sign language videos.
\newblock In {\em CVPR}, 2021.

\bibitem{Vaswani2017}
Ashish Vaswani, Noam Shazeer, Niki Parmar, Jakob Uszkoreit, Llion Jones, Aidan~N Gomez, \L~ukasz Kaiser, and Illia Polosukhin.
\newblock Attention is all you need.
\newblock In I. Guyon, U.~Von Luxburg, S. Bengio, H. Wallach, R. Fergus, S. Vishwanathan, and R. Garnett, editors, {\em Advances in Neural Information Processing Systems}, volume~30. Curran Associates, Inc., 2017.

\bibitem{yamada2018paired}
Tatsuro Yamada, Hiroyuki Matsunaga, and Tetsuya Ogata.
\newblock Paired recurrent autoencoders for bidirectional translation between robot actions and linguistic descriptions.
\newblock {\em IEEE Robotics and Automation Letters}, 3:3441--3448, 10 2018.

\bibitem{Zhang2023}
Jianrong Zhang, Yangsong Zhang, Xiaodong Cun, Shaoli Huang, Yong Zhang, Hongwei Zhao, Hongtao Lu, and Xi Shen.
\newblock T2m-gpt: Generating human motion from textual descriptions with discrete representations.
\newblock In {\em Proceedings of the IEEE/CVF Conference on Computer Vision and Pattern Recognition (CVPR)}, 2023.

\end{thebibliography}
}

\appendix
% \appendixpage
% \renewcommand{\thesubsection}{\Alph{subsection}}

\renewcommand{\thesection}{\Alph{section}}
\renewcommand{\thesubsection}{\thesection.\arabic{subsection}}

\section*{Supplementary}

\section{Introduction}

The motivation of our work stems from our additional goal of synchronizing text with motion using motion captioning as an intermediate task due to its applications such as aligned sign language transcription, unsupervised action segmentation and human motion segmentation. Unsupervised synchronization of text with motion requires a specific focus in architecture design that will be further analyzed in this supplementary material. In addition to animations, this supplementary feature introduces static visualizations as a preliminary to showcase the alignment between attention time for motion words and the corresponding retrieved primitives. Subsequently, we present additional quantitative results, followed by the illustration of qualitative assessments through static visualizations.

\section{Ablation analysis}
\label{supp:ablation}

In addition to the results mentioned in the paper, we highlight the following other important analysis.

\subsection{Multilayer vs. 1-layer Transformer}

\begin{figure*}[ht]
  \centering
  \includegraphics[width=\linewidth]{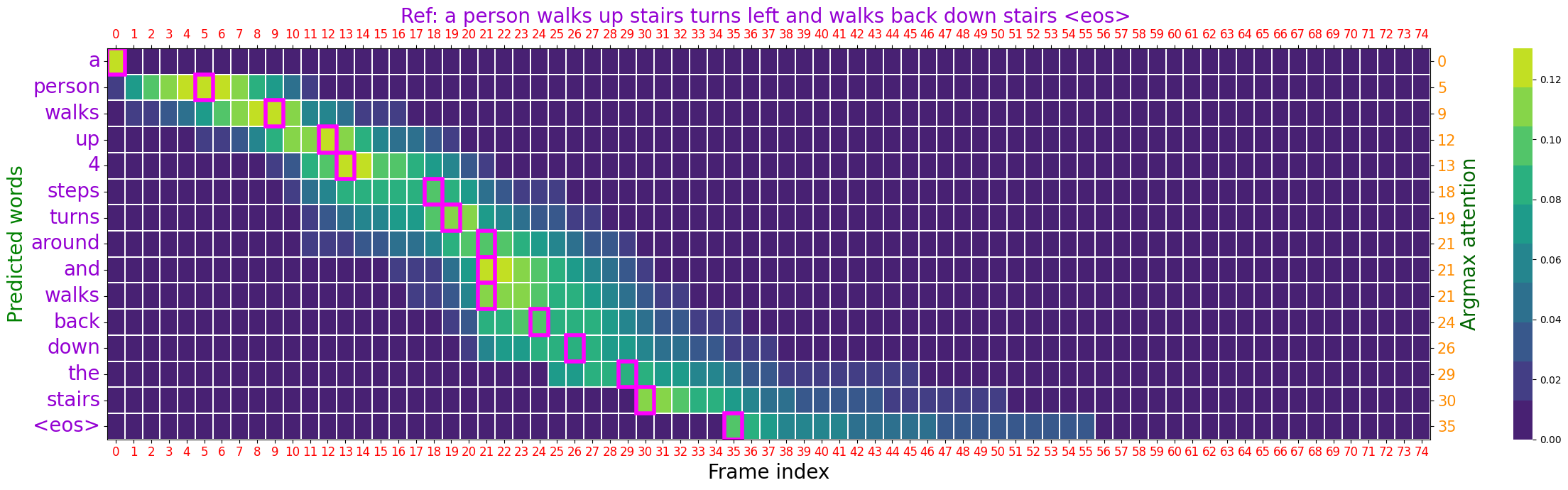}
   \caption{\textbf{Multi-Layer Transformer}: Compared to Figure \ref{fig:walk_turn_walk} attention distributions, here, are uninformative about action times, attention weights (for \textit{'turns'}, \textit{'walks back down'}) are not aligned with action times (same observation for different samples).}
   \label{fig:multi_layer_impact}
\end{figure*}

To demonstrate the sufficiency of our 1-layer based Transformer design, we compare the results against a multilayer transformer with 3 layers in both the encoder and decoder. The quantitative effect is discussed in Table \ref{tab:single_vs_multi_layer}. Qualitative impact is shown in Figure \ref{fig:multi_layer_impact}. The multilayer setting did not enhance text quality or synchronization performance beyond not being lightweight.

\begin{table}[h]
    \centering
    % \resizebox{.99\linewidth}{!}{
    \begin{tabular}{c|c|c|c|c|c}
        \# Layers & Mask.  & BLEU@4 $\uparrow$ &IoU $\uparrow$ & IoP $\uparrow$  & Element of $\uparrow$ \\ \hline
         \multirow{2}{*}{\textbf{1}} & No  & 26.5 & 39.93 & 39.95  & 48.98\\
         & \textbf{Yes}  & \textbf{27.1}  & \textbf{51.35}& \textbf{60.55} & \textbf{\textcolor{green}{71.55}} \\ \hline
        \multirow{2}{*}{3}  & No  & 25.7 & 41.88 & 41.92  & 39.81\\                   
                 & Yes  & 25.9 & 45.16 & 55.60 & \textcolor{red}{49.06}\\ 
      
    \end{tabular}
    
%    }
    \caption{\textbf{Even with Masking in Multi-layer Transformer}, the synchronization scores remain low compared to a single layer. This occurs because the \textbf{receptive field increases} across layers, causing \textbf{mixing information} in frame representations at the \textbf{final} encoder layer where representations of \textbf{early} frames contains information about very distant frames, leading to attention concentration at the beginning (See Figure \ref{fig:multi_layer_impact}). }
    \label{tab:single_vs_multi_layer}
\end{table}

\begin{figure*}[ht]
  \centering
  \includegraphics[width=\linewidth]{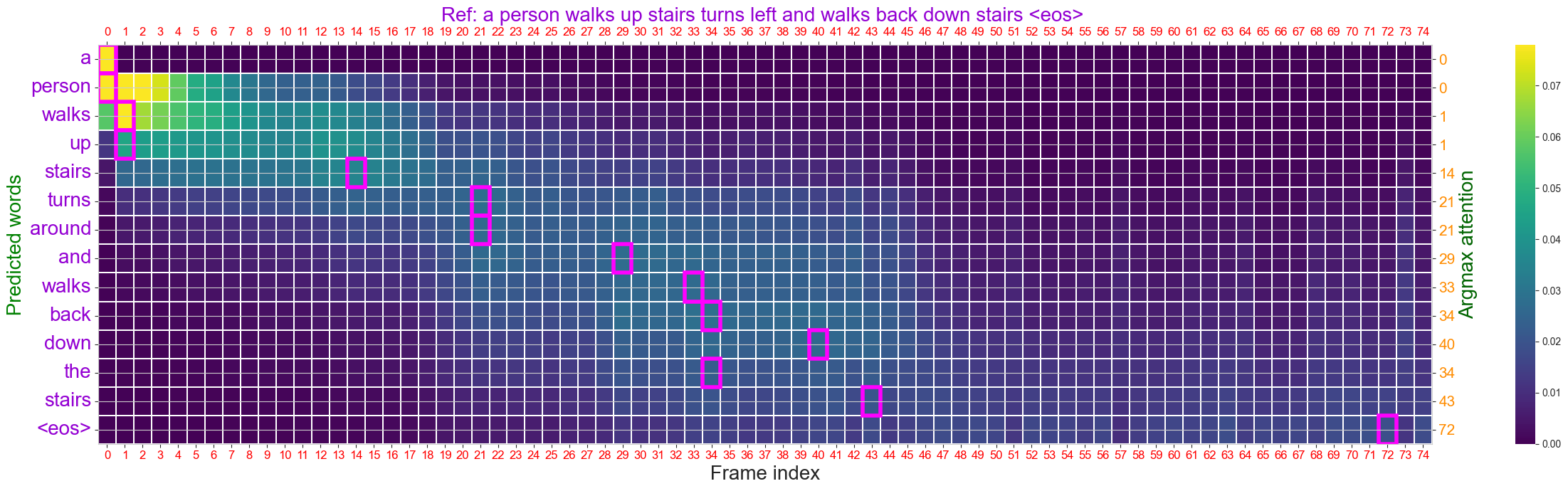}
\caption{\textbf{1-Layer Without Masking}: Compared to Figure \ref{fig:walk_turn_walk} \textit{'walks'} word attention is maximal around frame $1$, while this action starts at $\bf 10$. \textit{'turns'} (frame $21$) vs. correct range $[41,59]$. \textit{'Walk down stair'} highlighted in range $[33,43]$ vs. $[60,74]$. Our masking strategies were crucial in solving these issues.}
   \label{fig:control_w/o_mask}
\end{figure*}

\subsection{Masking and Attention Controlling Impact}

\textbf{Without masking} (See Figure \ref{fig:control_w/o_mask}) attention weights are uninformative about action times, this highly impacts synchronization scores (See Table \ref{tab:single_vs_multi_layer}, case 1-layer) which demonstrates the importance of our masking strategies.

\begin{figure*}[t]
  \centering
  \includegraphics[width=\linewidth]{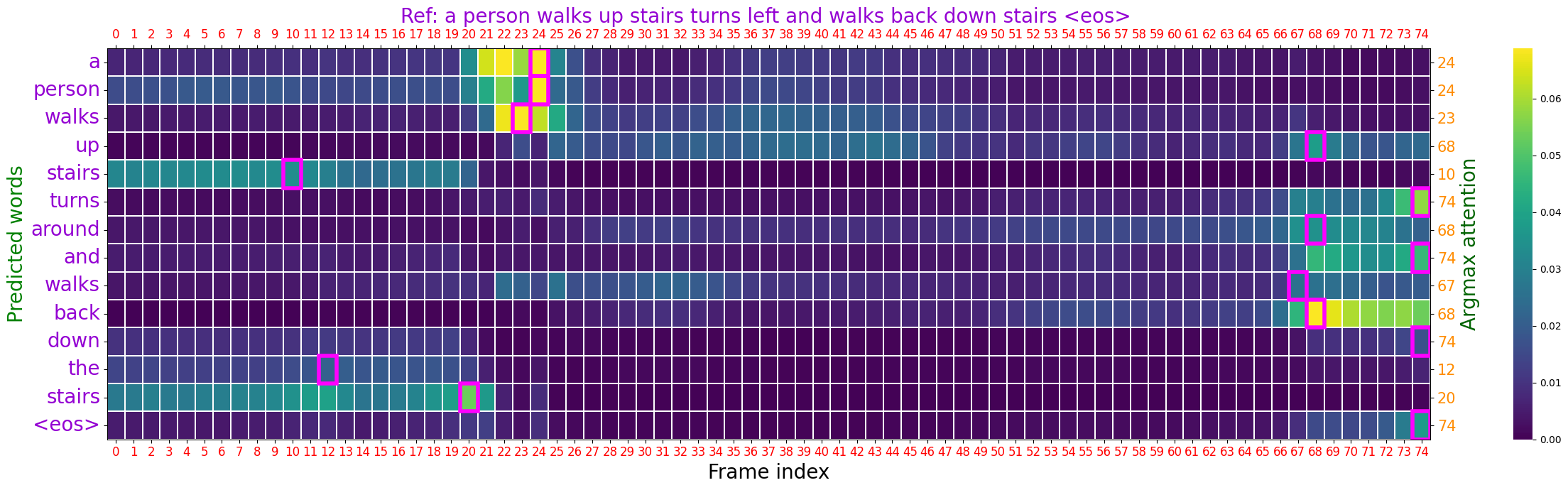}
   \caption{\textbf{Without Attention Control and Masking}: attention distributions are disordered and not very informative about action times (the same observation holds for different samples). Our masking and attention control were crucial in solving these issues (Cf.Tab \ref{tab:single_vs_multi_layer}-case 1-Layer).}
   \label{fig:nocontrol_and_mask}
\end{figure*}

\indent \textbf{Without Attention Control and Masking} attention weights don't carry any information about the action's timing or the order of execution (See Figure \ref{fig:nocontrol_and_mask}).

\FloatBarrier

\begin{figure*}[ht]
  \centering
  \includegraphics[width=\linewidth]{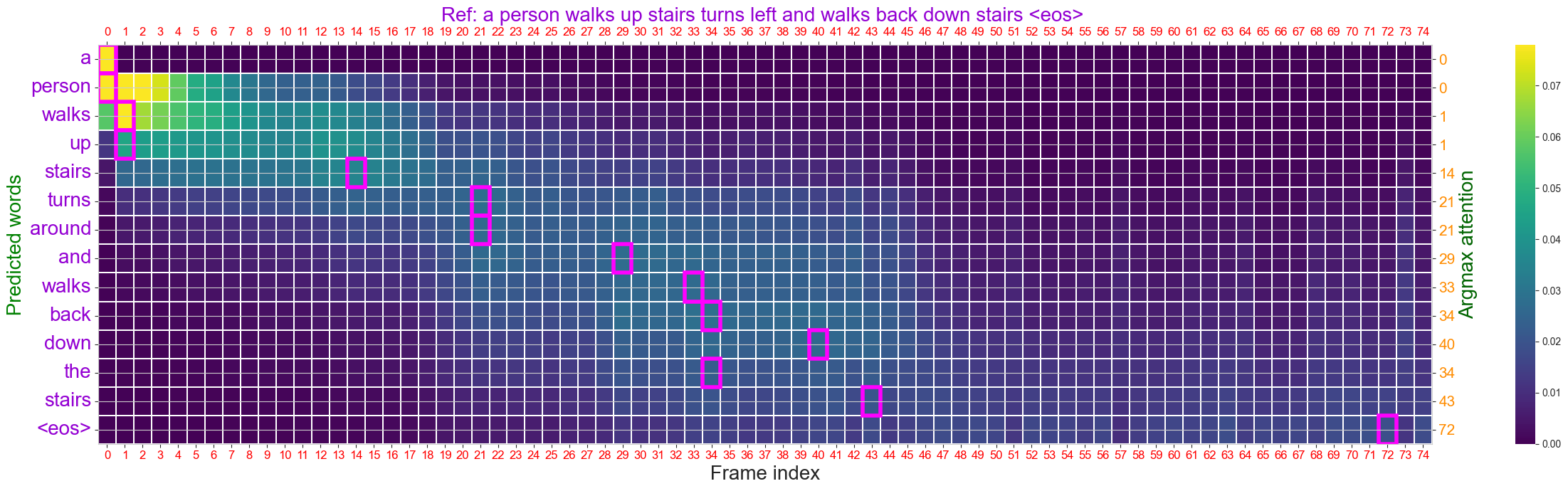}
\caption{\textbf{Controlling Attention without Masking}: \textit{walks} word attention is maximal around frame $1$, while this action starts at $\bf 10$. \textit{turns} (frame $21$) vs. correct range $[41,59]$. \textit{Walk down stair} highlighted in range $[33,43]$ vs. $[60,74]$. These issues occurs because early frames have access to all distant frames without masking.}
   \label{fig:control_w/o_mask}
\end{figure*}

\FloatBarrier
\section{Visualizations}
\label{supp:more_vizu}
Our Controlled and Masked Transformer is designed to enable action localization solely through attention. The current synchronization involves \textit{word-events}, but words describing the same event (action) could also be grouped, with attention weights aggregated by averaging across relevant language segments to form \textit{phrase-events} association. In this part, we provide static visualizations with motion frozen in time, given by the key attention frames (motion frames receiving maximum attention) at word- and phrase-level. Then, we visualize some additional cross-attention maps associating human motion sequences and language word descriptions in time.

\subsection{Motion Frozen in Time}

 In this section, we aim to illustrates poses sequence and motion words association based on attention weights. We present static visualizations capturing motion frozen in time accompanied by corresponding descriptive words. Nevertheless, as previously explained, static visualizations inherently possess limitations, making them a complement to animations.

\subsubsection{Word level attention}
For each word, we visualize \textit{four motion frames receiving maximum attention at inference time.}

\begin{figure*}[h]
    \centering
    \includegraphics[width=0.8\columnwidth]{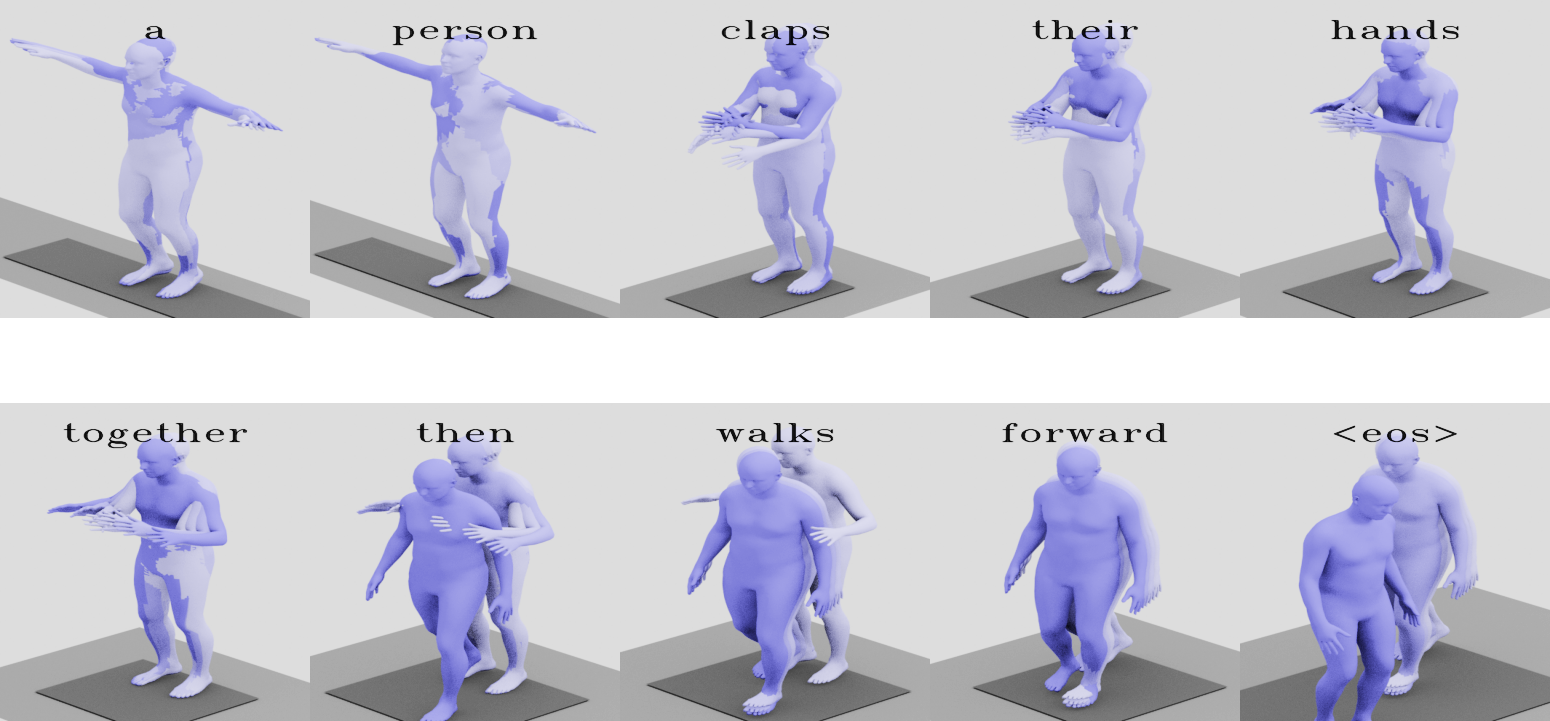}
    \caption{a person \textbf{claps} their \textbf{hands} then \textbf{walks forward}.}
    \label{fig:enter-label}
\end{figure*}

\begin{figure*}[h]
    \centering
    \includegraphics[width=0.9\columnwidth]{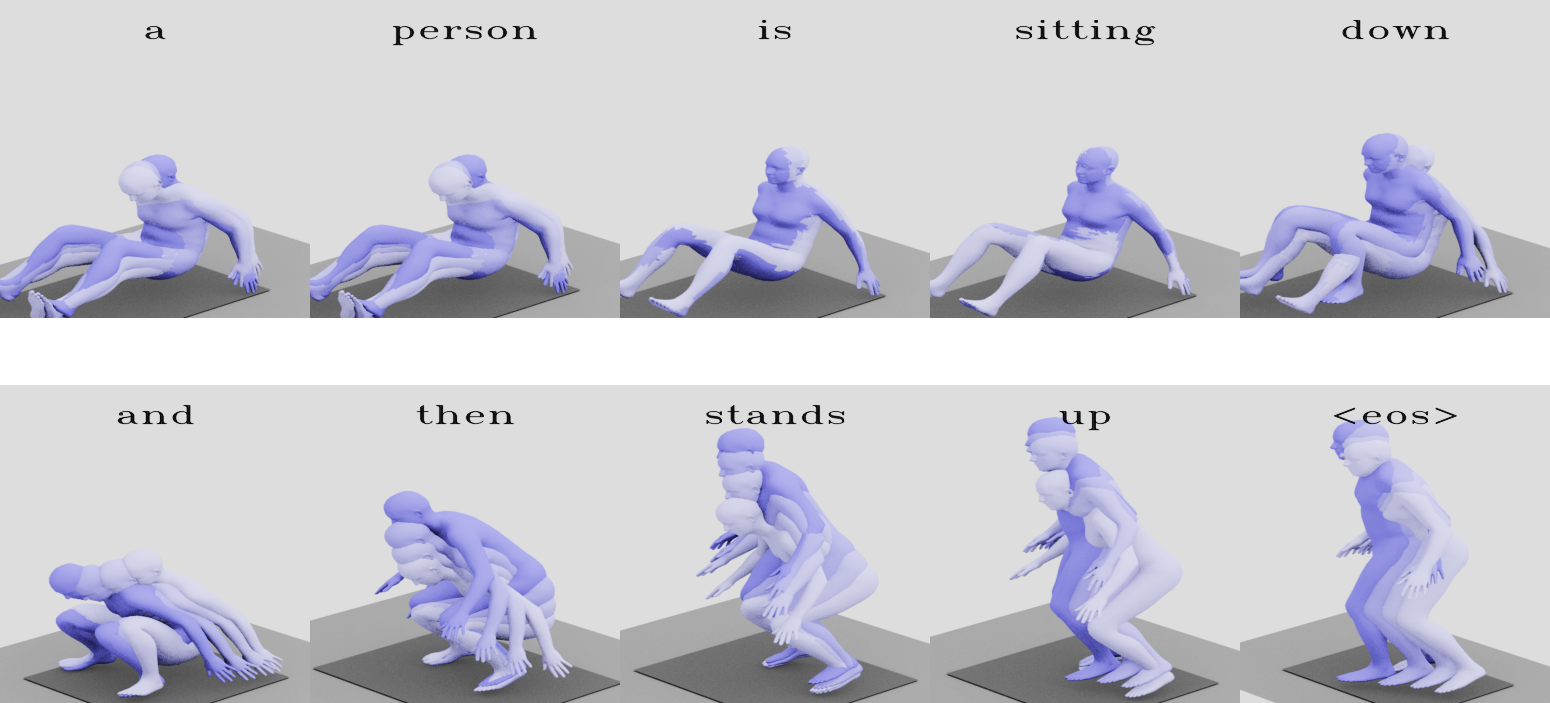}
    \caption{ a person is \textbf{sitting down} and then \textbf{stands up}.}
    \label{fig:enter-label}
\end{figure*}

\begin{figure*}[h]
    \centering
    \includegraphics[width=0.9\columnwidth]{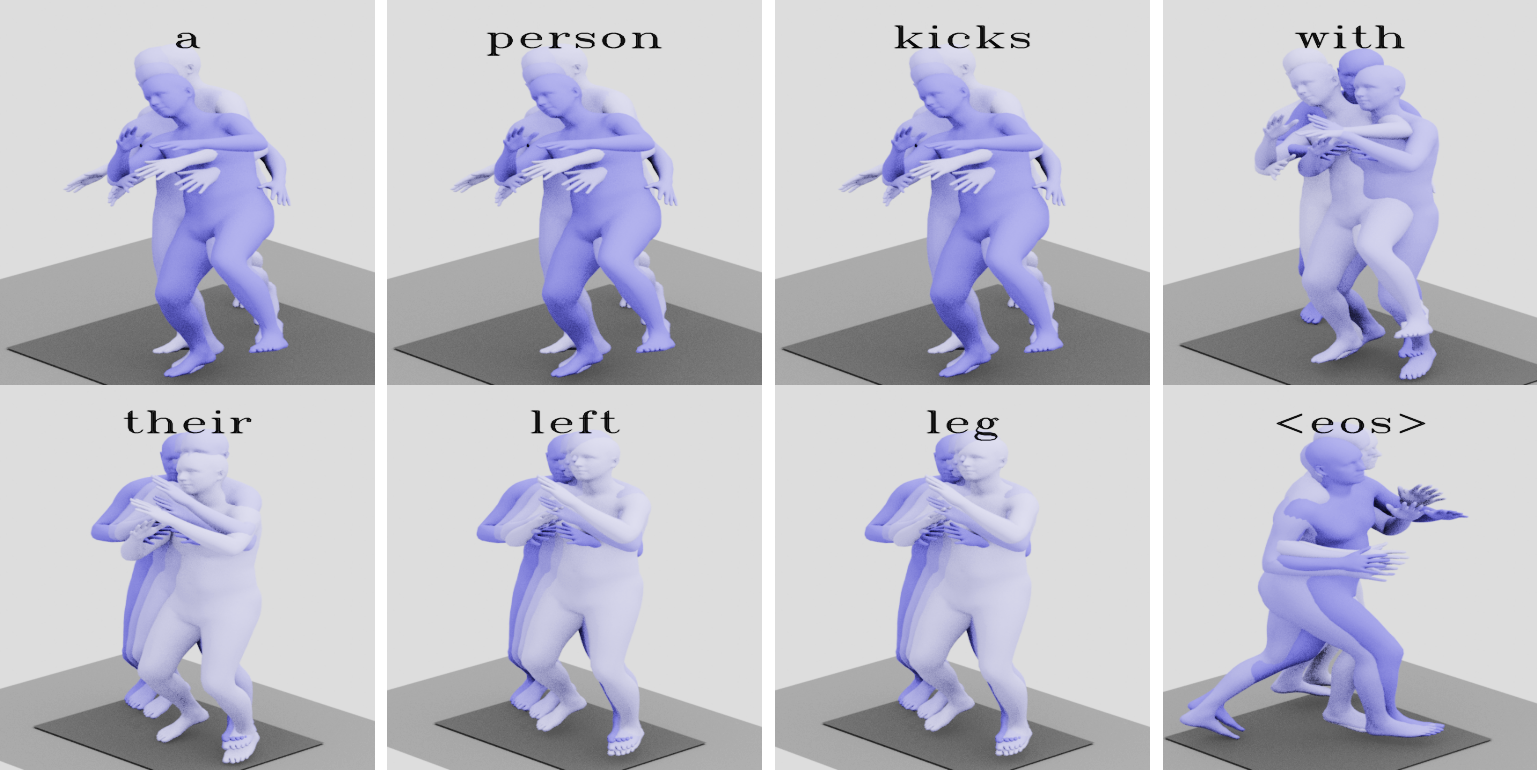}
    \caption{ a person \textbf{kicks} with their \textbf{left leg}.}
    \label{fig:enter-label}
\end{figure*}

\begin{figure*}[h]
    \centering
    \includegraphics[width=0.9\columnwidth,height=10cm]{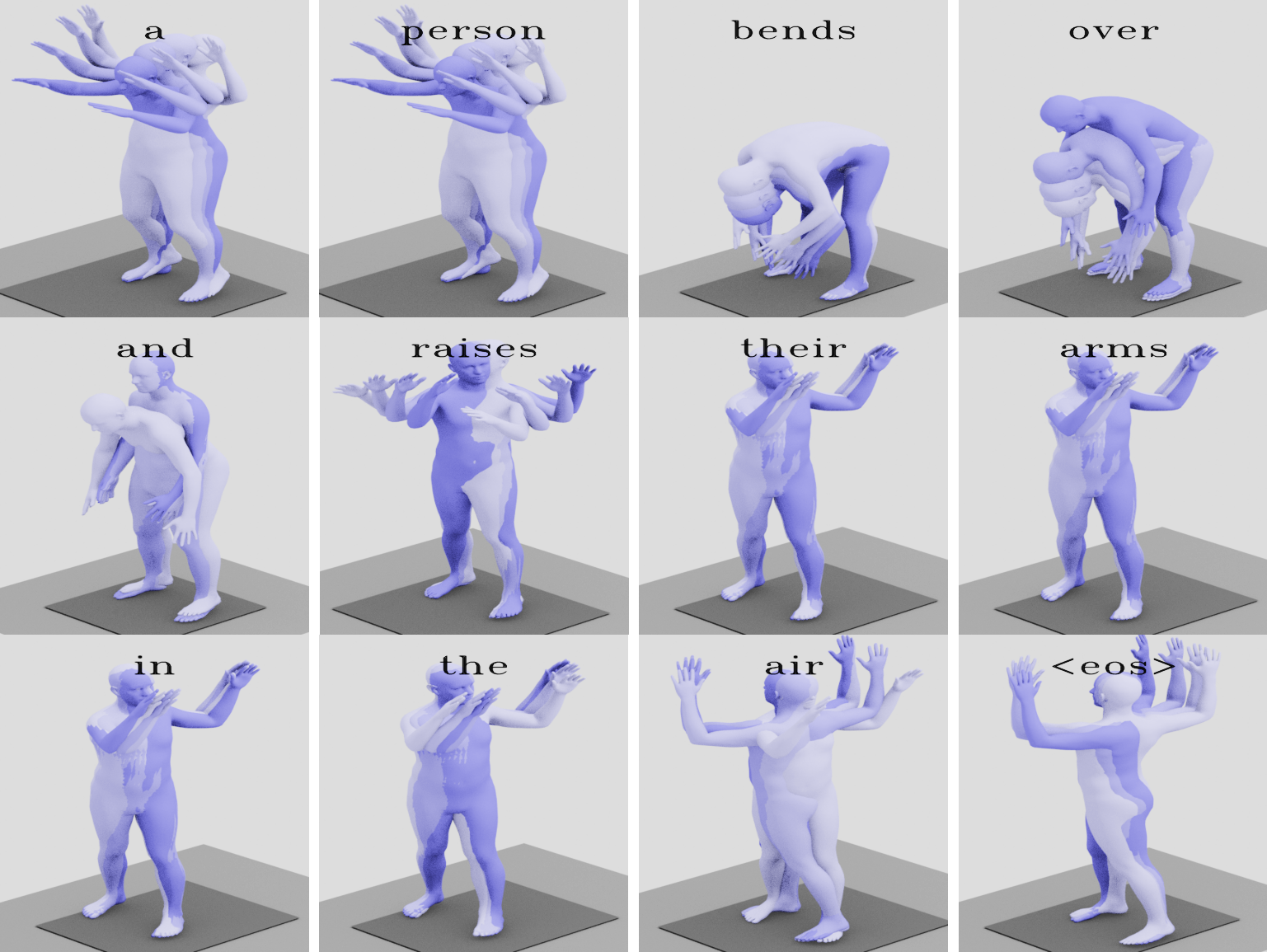}
    \caption{ a person \textbf{bends} over and \textbf{raises} their \textbf{arms} in the air.}
    \label{fig:enter-label}
\end{figure*}

\begin{figure*}[h]
    \centering
    \includegraphics[width=0.9\columnwidth,height=10cm]{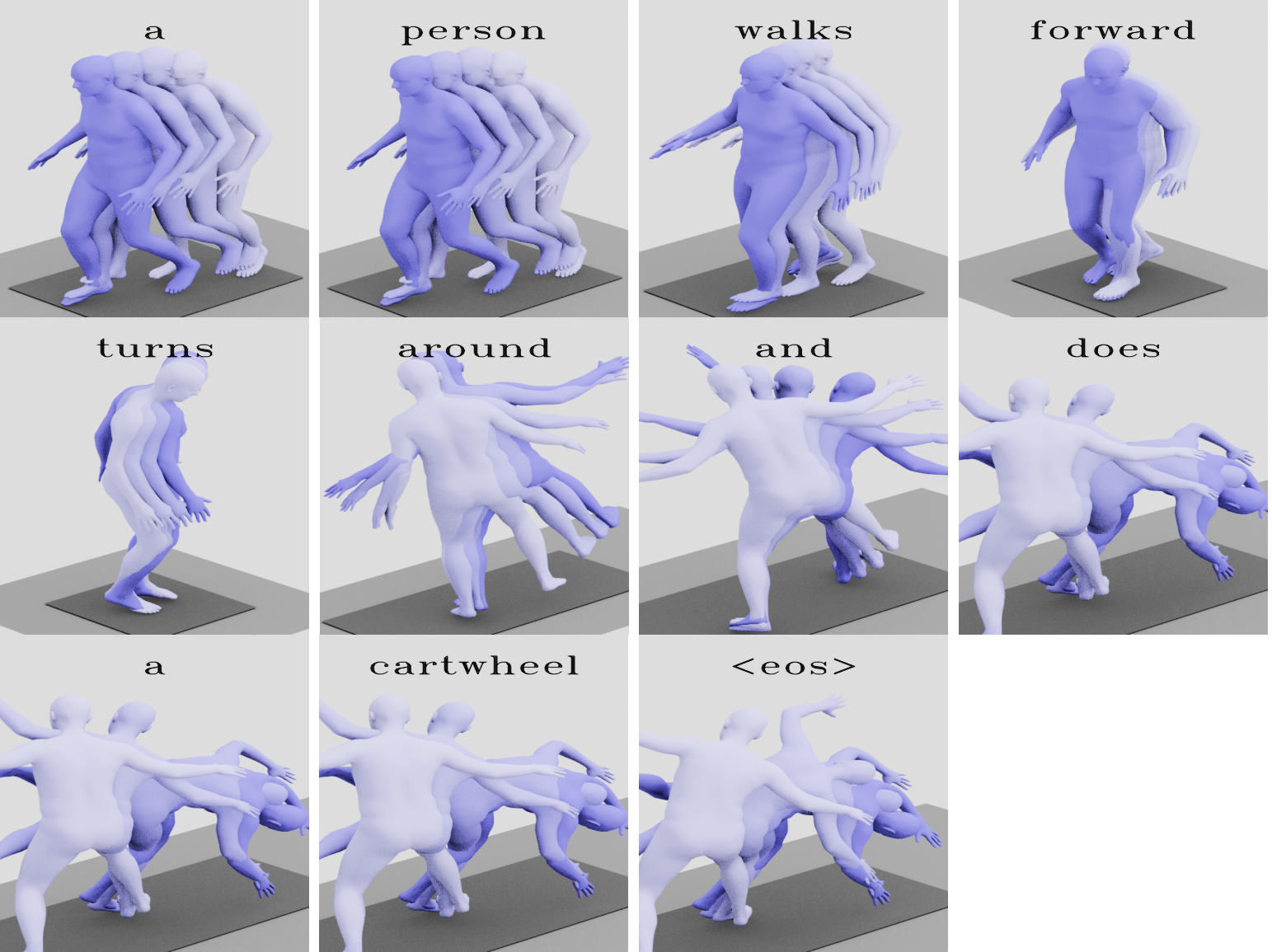}
    \caption{a person \textbf{walks forward turns around} and does a \textbf{cartwheel}.}
    \label{fig:enter-label}
\end{figure*}

\FloatBarrier
\subsubsection{Phrase level attention}
 The attention weights are aggregated by averaging across words relative to the  primitive motion between motion words, then four frames of higher attention are displayed for each corresponding language segment.

\begin{figure*}[h]
    \centering
        \includegraphics[width = 0.9\textwidth,height=6cm]{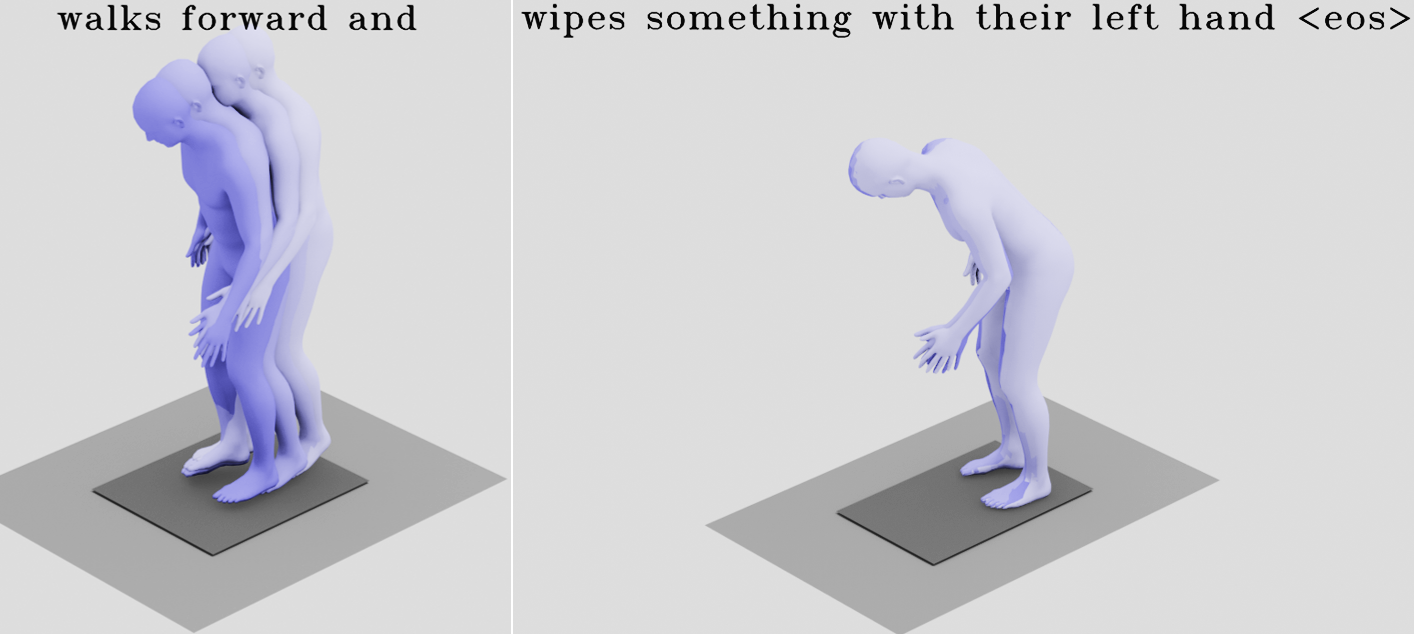}
        \includegraphics[width = 0.9\textwidth,height=6cm]{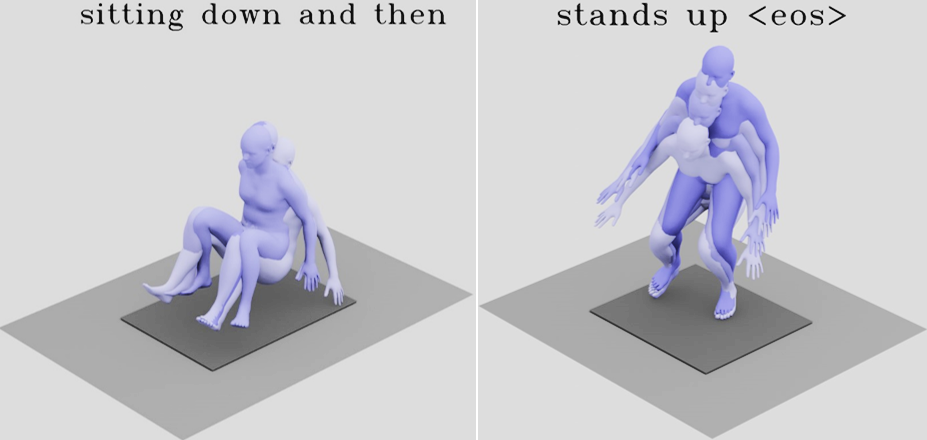}
        \includegraphics[width = 0.9\textwidth,height=6cm]{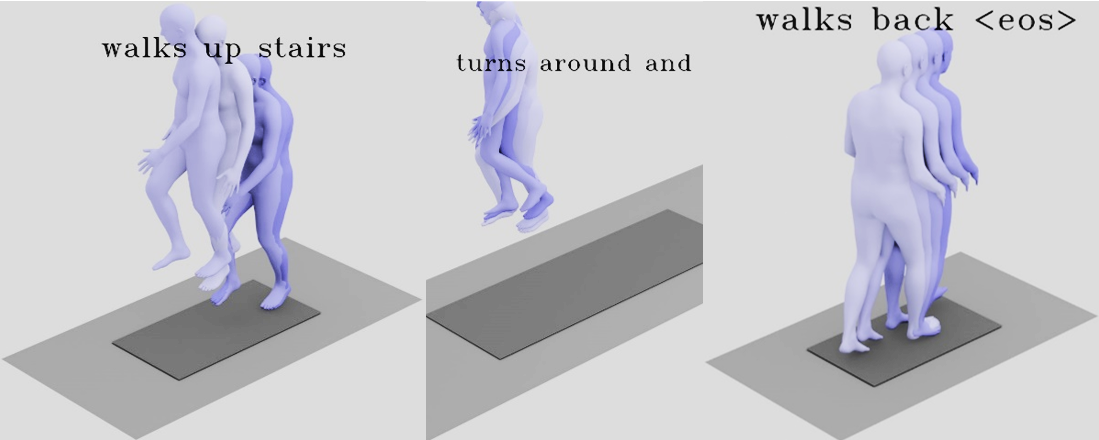}
        \caption{\textbf{Phrase-level} attention based association between motion and language segments.} 
        \label{fig:subfig1}

\end{figure*}

\FloatBarrier

\subsection{Cross attention maps}
In the following attention maps of different motions (some from the same samples visualized with frozen mesh above and in the paper above for correspondence):

\begin{figure*}[h]
    \begin{subfigure}{\textwidth}
        \includegraphics[width=\textwidth]{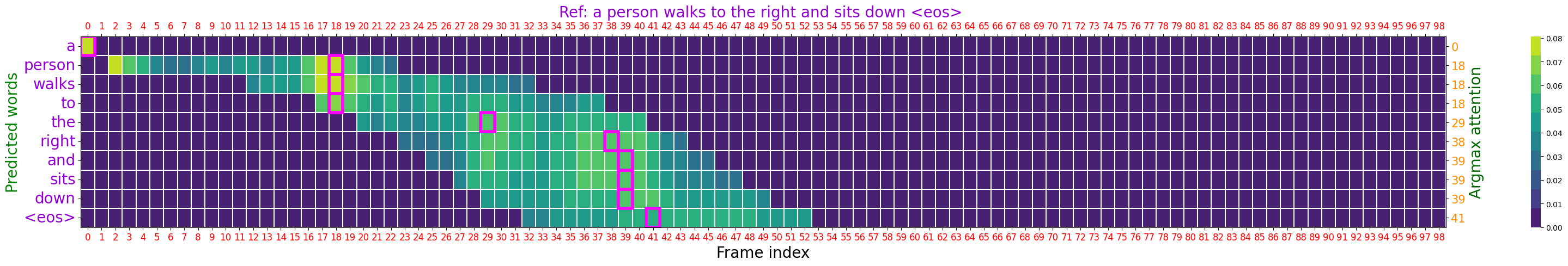}
        \caption{Walks to the right, sits down.}
        \label{fig:subfig2}
    \end{subfigure} \vfill
        \begin{subfigure}{\textwidth}
        \includegraphics[width=\textwidth]{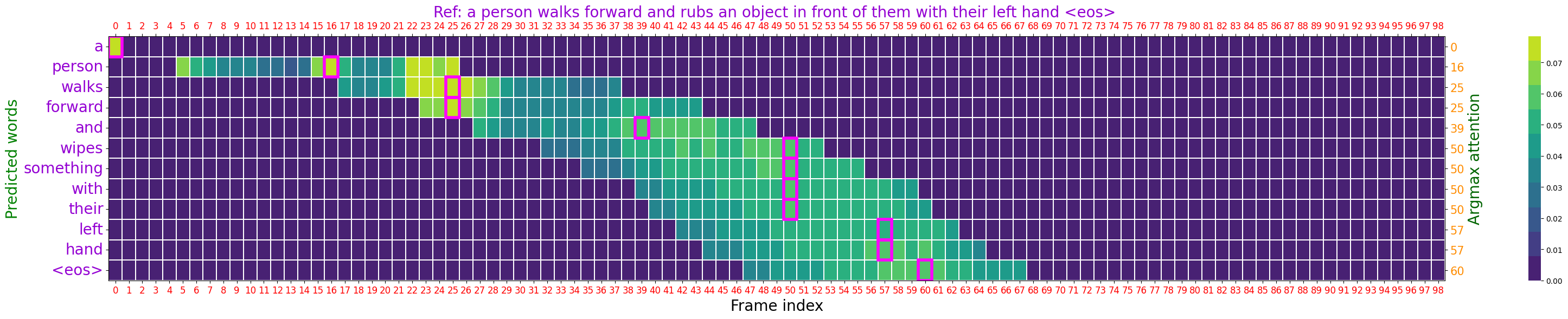}
        \caption{Walks forward, wipes something.}
        \label{fig:subfig2}
    \end{subfigure} 
\end{figure*}

\begin{figure*}[h]
    \begin{subfigure}{\textwidth}
        \includegraphics[width=\textwidth]{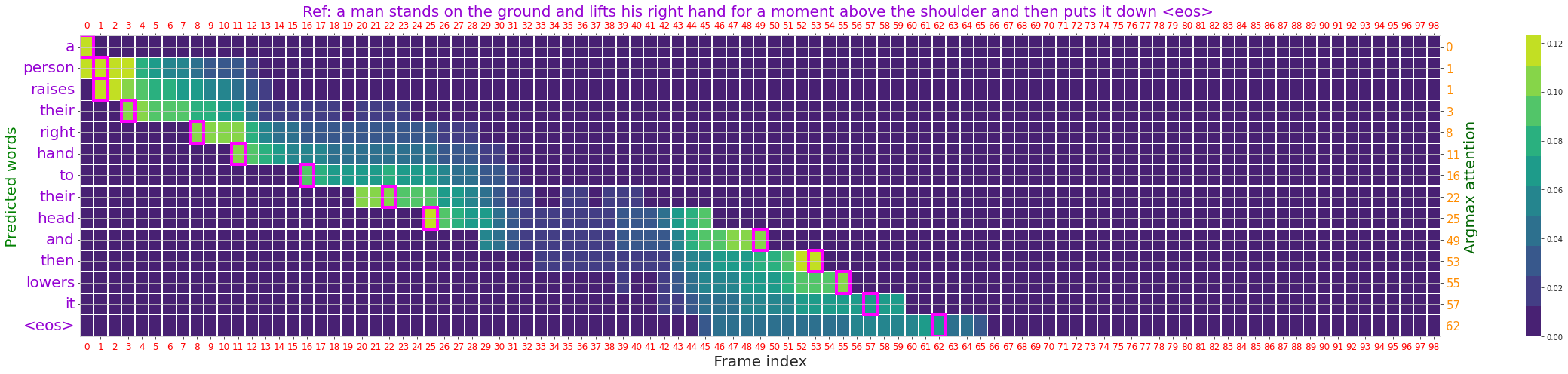}
        \caption{raise right hand to shoulder/head level then put it down (right hand).}
        \label{fig:subfig2}
    \end{subfigure} \vfill
        \begin{subfigure}{\textwidth}
        \includegraphics[width=\textwidth]{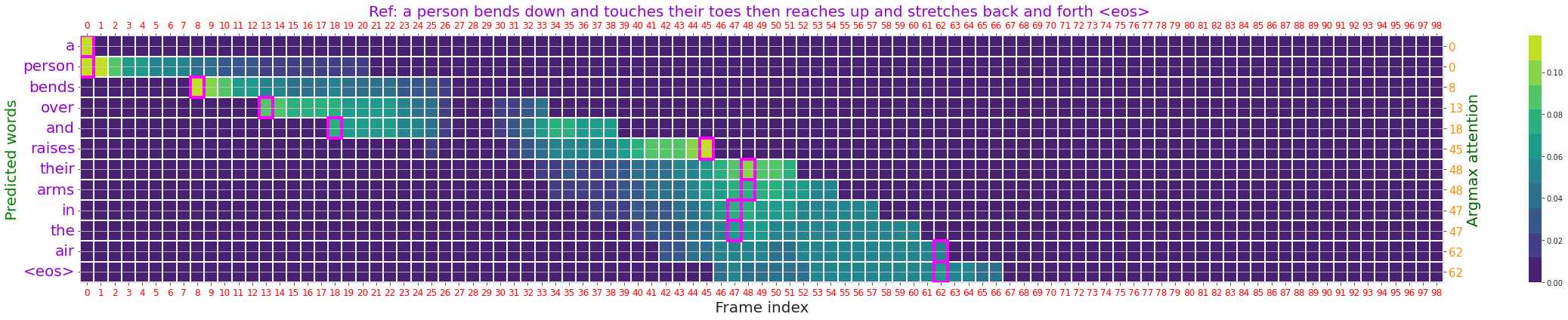}
        \caption{bends over and raises arms in the air.}
        \label{fig:subfig2}
    \end{subfigure} 
\end{figure*}

\begin{figure*}[h]
    \centering
    \begin{subfigure}{\textwidth}
        \includegraphics[width = \textwidth]{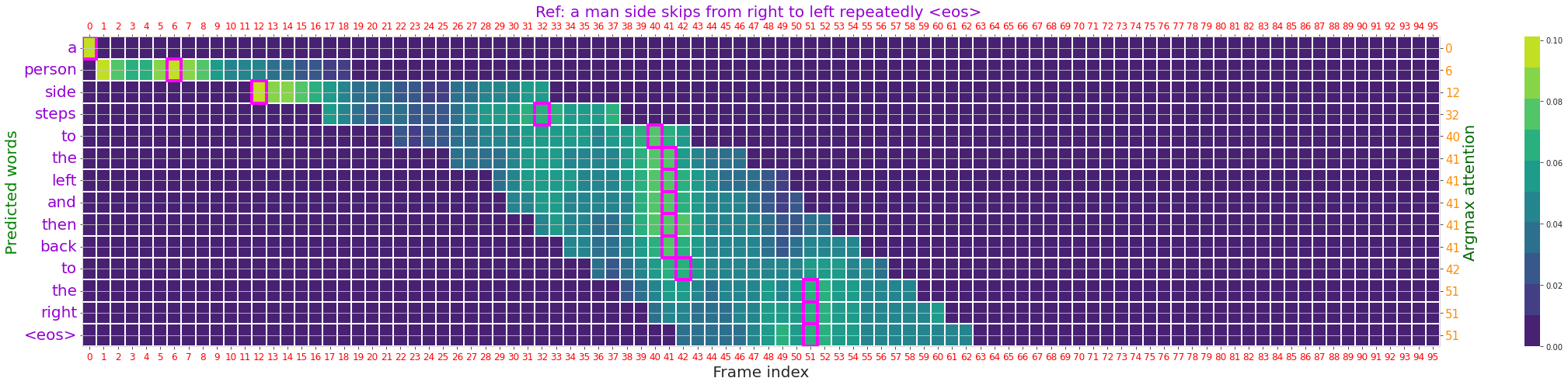}
        \caption{Side step to the left then to the right.}
        \label{fig:subfig1}
    \end{subfigure} \vfill
    \begin{subfigure}{\textwidth}
        \includegraphics[width=\textwidth]{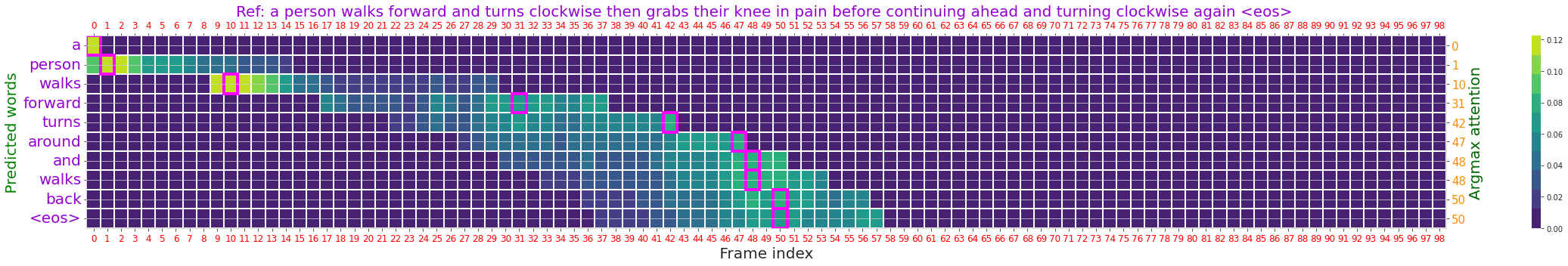}
        \caption{walks forward, turn around then walks back.}
        \label{fig:subfig2}
    \end{subfigure} \vfill
\end{figure*}

\begin{figure*}[h]
    \begin{subfigure}{\textwidth}
        \includegraphics[width=\textwidth]{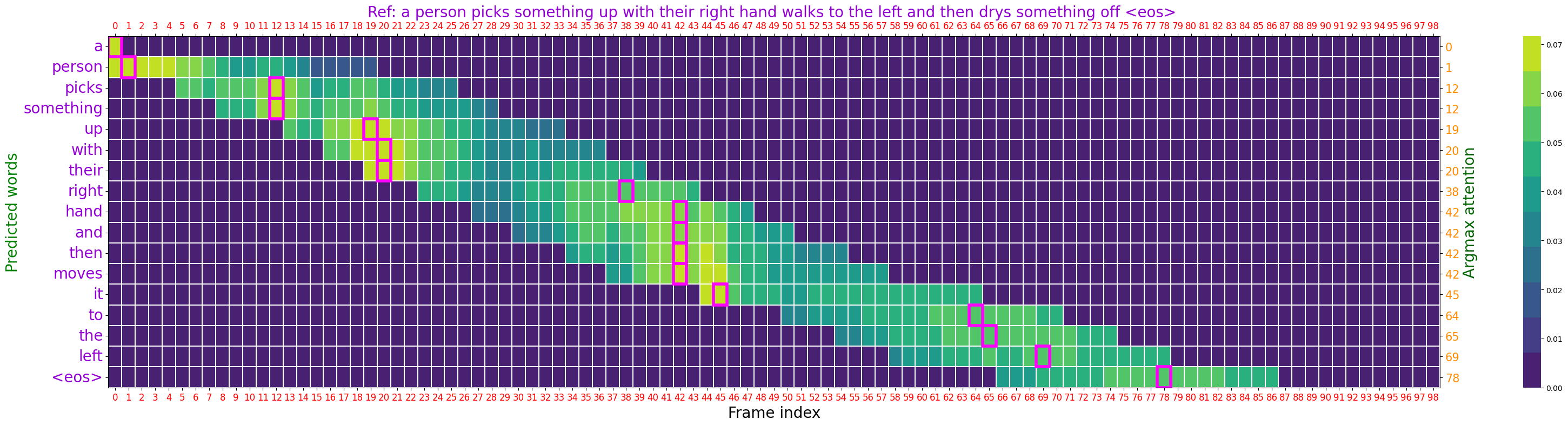}
        \caption{picks something with right hand, move it to left.}
        \label{fig:subfig2}
    \end{subfigure} \vfill
        \begin{subfigure}{\textwidth}
        \includegraphics[width=\textwidth]{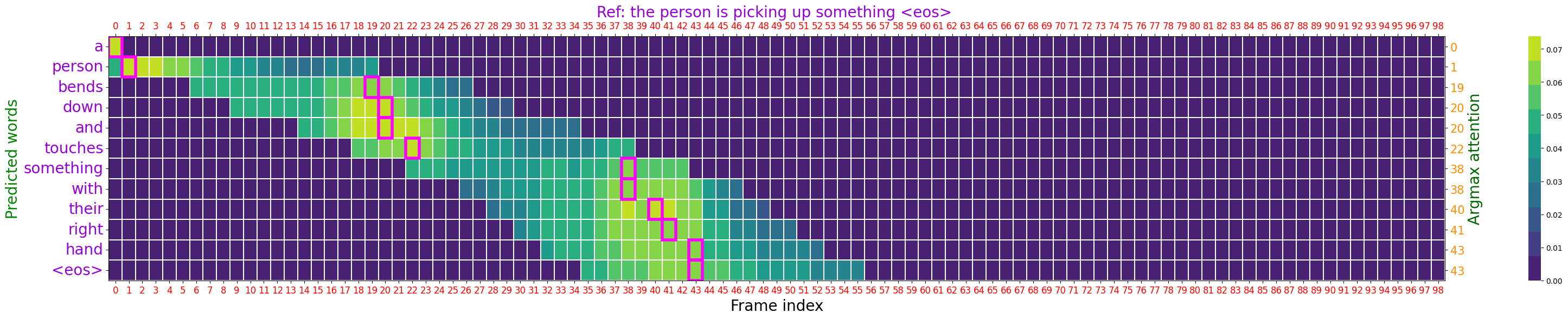}
        \caption{bends down and touches something.}
        \label{fig:subfig2}
    \end{subfigure} 
\end{figure*}

\end{document}